# A Finite Markov Chain Analysis of the Simplified Swarm Optimisation Update Rule

Wei-Chang Yeh[a,*]

[a]*Department of Industrial Engineering and Engineering Management, National Tsing Hua University, No. 101, Section 2, Kuang-Fu Road, Hsinchu 30013, Taiwan*

[*]Corresponding author. E-mail address: yeh@ieee.org

**Abstract**—Simplified swarm optimisation (SSO) updates each variable by one draw that copies the global or personal best, keeps the current value or redraws uniformly, with probabilities $c_g$, $c_p$, $c_w$ and $c_r$. As a finite Markov chain under frozen memory, its coordinate kernel $c_wI + (1 - c_w)\Pi$ has an explicit stationary law and exact convergence rate $c_w^t$. The elitist chain has a geometric hitting-time bound, from a non-optimal start a fixed $c_r$ needs at least $\exp(nc_r/2)$ expected evaluations to reach a unique optimum, and a coupling under changing memories gives $O(n \log n)$ and $O(n^2)$ expected generations on OneMax and LeadingOnes for every population size with redraw share of order 1/n and no inertia, and within a log n factor for fixed inertia. With a frozen memory, inertia cannot speed up first entry into any set from stationary positions, though it can from others. Where the three reference values agree, only $c_r$ acts; where they disagree, copy branches recombine the memories, forcing a redraw only where no active branch holds the target value. Branch logging and checkpoint replay show inertia slowed the tested OneMax and LeadingOnes runs and suggest memory-mediated gains on wider gaps. On Jump functions SSO needed 3.8–8.9 times fewer evaluations than the (1+1) evolutionary algorithm with rate 1/n, but no consistent advantage over rate-adjusted mutation or a (μ+1) genetic algorithm with uniform crossover was established. The analytical redraw rule improved six of eight benchmark–setting pairs and redundancy allocation, but was slower than $c_r = 0.1$ on $\mathrm{Jump}_2$.

## I. Introduction

Simplified swarm optimisation (SSO) was introduced by Yeh [1] as a discrete swarm method in which every variable is updated by one comparison of a uniform random number against a small set of cumulative thresholds. The rule has no velocity, no inertia vector and no learning-factor product, which makes it easy to implement and has led to many applications, including reliability redundancy allocation [1]–[3], disassembly sequencing [4], data clustering [5] and the hyperparameter tuning of convolutional neural networks [6]. Its theory has lagged behind its use.

Two lines of theory are relevant. Markov chain arguments establish global convergence of elitist evolutionary algorithms: Rudolph [7] showed that an elitist genetic algorithm reaches the optimum with probability one because its mutation gives every point positive probability, and Theorems 2–3 below are the SSO counterpart of that argument. Runtime analysis then bounds the expected time, typically by fitness levels [8] or drift [9], and these methods have been carried over to swarm algorithms on discrete spaces. Sudholt and Witt [10] adapted the fitness-level method to a binary particle swarm optimiser and analysed it on OneMax and other unimodal functions, and Mühlenthaler et al. [11] analysed a particle swarm optimiser for discrete problems through a Markov model, using indistinguishable states to obtain upper and lower bounds on sorting and OneMax that are essentially tight with respect to that model. Trajectory and stability analyses of continuous particle swarm optimisation [12], [13] address a different question and are not used here. The SSO kernel differs from both discrete swarm algorithms: it is a coordinatewise mixture of copying from three sources and a uniform redraw, with no velocity and no attraction probabilities. This product structure is what yields a closed-form frozen-memory spectrum for every population size, a transfer from one solution to the swarm through the owner of the global best while the memories change, and an exact comparison between inertia and independent sampling. Population diversity and crossover are known to help on functions with fitness gaps: the (μ+1) GA gains from crossover with emergent diversity [14], and a strong drift towards diverse populations gives it $O(n^{k-1})$ expected evaluations on $\mathrm{Jump}_k$ for $k \geq 3$ [15]. Section VIII-B identifies a related recombination effect inside the SSO update, and Section X compares the two. The contributions fall into three groups:

1) **Exact structure of the update kernel.** A closed form of the frozen-memory coordinate kernel, its stationary law and its exact decay rate $c_w^t$ (Theorem 1); state classification, global convergence and a geometric hitting-time bound whose one-step ingredient is sharp, and a lower bound of $\exp(nc_r/2)$ expected evaluations for every redraw weight held fixed during the run, when the optimum is unique and the initial global best is not optimal (Theorems 2–3, Proposition 2); a gBest-guided bound with the end-game redraw weight $m/((m-1)n)$ (Theorem 4, Corollary 1); and the effective branch probabilities when reference values coincide (Proposition 4). Section V treats a complement-mutation variant.
2) **Runtime under changing memories.** Using the owner of the global best, an explicit coupling and the exact decay rate, expected-runtime bounds on OneMax and LeadingOnes for every population size and every fixed inertia weight, with the redraw share scaled as $m/((m-1)n)$ (Section VI).
3) **Inertia and recombination.** A comparison of inertia with independent sampling for a frozen memory and stationary positions (Theorem 8), with an exact example showing that it does not extend to non-stationary positions (Example 1); a recombination bound that forces a redraw only where no active branch holds the target value, with its consequence for complementary local optima of $\mathrm{Jump}_k$ (Proposition 5, Corollary 4); and experiments that separate these mechanisms by branch-level logging, checkpoint replay and mechanism-matched controls (Sections IX-F to X-B).

Sections IX and X validate the results by exact computation and simulation and test them on benchmark functions and on the 33 instances of the Fyffe–Hines–Lee redundancy allocation problem, with exact optima as reference.

For a single solution without inertia on bits, SSO reduces to the (1+1) evolutionary algorithm (EA) with strict acceptance and bit-flip probability $c_r/2$, so the runtime results then reproduce known bounds [16], [17]. This case serves as a consistency check; the new content lies in the general memory structure and in the role of inertia. The continuous version of SSO, in which the three copy branches become Gaussian perturbations, requires general state-space arguments and is treated in a companion paper [18].

## II. The SSO Update Rule as a Markov Chain

A population of N solutions $X = (x_1, \ldots, x_N)$ is searched over $D = S_1 \times \ldots \times S_n$ with $|S_j| = m_j$ and $2 \leq m_j < \infty$ for every j; we write m when all $m_j$ are equal, and $n \geq 2$; in asymptotic statements $n \to \infty$, while m and, where stated, the inertia weight $c_w$ are fixed and the redraw share is scaled with n as stated. Each solution i keeps a personal best $p_i$; the global best g is the best personal best. The objective f is minimised and $D^* = \arg\min f$. The thresholds satisfy $0 \leq C_g \leq C_p \leq C_w \leq 1$, and the lower-case quantities $c_g = C_g$, $c_p = C_p - C_g$, $c_w = C_w - C_p$ and $c_r = 1 - C_w$ are the probabilities of the four branches (copy gBest, copy pBest, keep the current value, random redraw), so $c_g + c_p + c_w + c_r = 1$. For each i and j, draw $\rho \sim U[0, 1)$ independently and set

$$x'_{ij} = g_j \text{ if } \rho < C_g;\ \ p_{ij} \text{ if } C_g \leq \rho < C_p;\ \ x_{ij} \text{ if } C_p \leq \rho < C_w;\ \ u \sim U(S_j) \text{ otherwise.} \tag{1}$$

After all N new solutions are produced (synchronous update), $p_i \leftarrow x'_i$ if $f(x'_i) < f(p_i)$, and g is reset to the best $p_i$, with ties broken by the smallest index. The state is $s = (X, P)$, since g is a function of P. The state space $\Omega = D^N \times D^N$ is finite and $s_t$ is a time-homogeneous Markov chain. We assume:

- **A1.** $C_w < 1$, that is $c_r > 0$.
- **A2.** Given $s_t$, the random numbers used for different variables and solutions are independent.
- **A3.** Personal bests are replaced only on strict improvement; g is the best personal best.

**Threshold conventions.** The SSO literature orders the four branches in more than one way. For example, [5] uses thresholds $C'_w \le C'_p \le C'_g$, with the current value for $\rho < C'_w$, the personal best for $C'_w \le \rho < C'_p$, the global best for $C'_p \le \rho < C'_g$ and a random value otherwise. Every result below depends only on the branch probabilities $c_g$, $c_p$, $c_w$ and $c_r$, which Table I converts between the two conventions; parameter values from earlier work should be converted before they are used in formulas such as $C_w - C_p$ or $1 - C_w$.

TABLE I. Branch probabilities under two threshold conventions

| Branch | Probability | This paper ($C_g \le C_p \le C_w$) | Convention of [5] ($C'_w \le C'_p \le C'_g$) |
|---|---|---|---|
| Copy gBest | $c_g$ | $C_g$ | $C'_g - C'_p$ |
| Copy pBest | $c_p$ | $C_p - C_g$ | $C'_p - C'_w$ |
| Keep current value | $c_w$ | $C_w - C_p$ | $C'_w$ |
| Random redraw | $c_r$ | $1 - C_w$ | $1 - C'_g$ |
| Example | (0.5, 0.3, 0.1, 0.1) | $(C_g, C_p, C_w) = (0.5, 0.8, 0.9)$ | $(C'_w, C'_p, C'_g) = (0.1, 0.4, 0.9)$ |

Algorithm 1 summarises one generation and Table II lists the notation. The analysis below needs only the kernel it induces.

**Algorithm 1** (one generation of discrete SSO).

1. For each solution $i = 1, \ldots, N$ and each variable $j = 1, \ldots, n$: draw $\rho \sim U[0, 1)$ and set $x'_{ij}$ by (1).
2. For each i: evaluate $f(x'_i)$; if $f(x'_i) < f(p_i)$, set $p_i \leftarrow x'_i$.
3. Set $i^* \leftarrow \arg\min_i(f(p_i), i)$, the smallest index whose personal best has the best fitness, and $g \leftarrow p_{i^*}$; set $X \leftarrow X'$.

TABLE II. Notation

| Symbol | Meaning |
|---|---|
| $C_g \le C_p \le C_w$ | Cumulative thresholds of SSO in the convention of this paper (Table I) |
| $c_g, c_p, c_w, c_r$ | Branch probabilities: copy gBest, copy pBest, keep current value, random redraw |
| $n, N, m_j$ | Number of variables, population size, number of values of variable j |
| $D, D^*$ | Search space $\prod_j S_j$ and set of global optima |
| $x_i, p_i, g$ | Current solution, personal best and global best |
| $s_t = (X_t, P_t)$ | Markov state at generation t |
| $T$ | First generation in which g is a global optimum |
| $q = c_r^n\|D^*\|/\|D\|$ | Worst-case one-step probability of sampling a global optimum |
| $i^*, r = c_r/(1 - c_w)$ | Owner of g (Lemma 2) and redraw share of the non-inertia mass (Section VI) |

## III. The Update Kernel and the Frozen-Memory Chain

**Proposition 1 (coordinate kernel).** *Under A2, for $y \in S_j$,*

$$K_j(y \mid x, p, g) = c_g \cdot 1\{y = g_j\} + c_p \cdot 1\{y = p_{ij}\} + c_w \cdot 1\{y = x_{ij}\} + c_r/m_j, \tag{2}$$

and the new population has probability $\prod_i \prod_j K_j(y_{ij} \mid x_i, p_i, g)$.

*Proof.* The four branches are disjoint events with probabilities $c_g$, $c_p$, $c_w$ and $c_r$; the uniform branch puts mass $c_r/m_j$ on every value. Independence is A2. ■

**Lemma 1 (Doeblin minorisation).** *For every state s and every $Y \in D^N$, $P(X_{t+1} = Y \mid s_t = s) \ge \alpha := (c_r^n/|D|)^N$.*

*Proof.* Each factor in Proposition 1 is at least $c_r/m_j$. ■

To isolate the role of the thresholds, fix P and g. The positions then evolve as a Markov chain on $D^N$ that factorises into nN independent coordinate chains. Let Q be the transition matrix of one coordinate on S = {1, …, m}, let $e_v$ be the unit vector of value v, let 𝟙 be the all-ones vector and let $\mu = c_g e_g + c_p e_p + (c_r/m)\mathbb{1}$, so $\mu^T\mathbb{1} = 1 - c_w$.

**Theorem 1 (closed form and exact decay).** *If $c_w < 1$, $Q = c_w I + \mathbb{1}\mu^T = c_w I + (1 - c_w)\Pi$ with $\Pi = \mathbb{1}\pi^T$ and $\pi = \mu/(1 - c_w)$. Hence (i) π is the unique stationary distribution:*

$$\pi(v) = [c_g\cdot 1\{v = g\} + c_p\cdot 1\{v = p\} + c_r/m] \,/\, (1 - C_w + C_p); \quad (3)$$

*(ii) the eigenvalues of Q are 1 (simple) and $c_w$ (multiplicity m − 1); (iii) $Q^t = \Pi + c_w^t(I - \Pi)$, so the total-variation distance from π started at x is exactly*

$$\|e_x^T Q^t - \pi^T\|_{TV} = (C_w - C_p)^t(1 - \pi(x)); \quad (4)$$

*(iv) for the whole population, $\|P^t(X, \cdot) - \otimes_{i=1}^N \otimes_{j=1}^n \pi_{ij}\|_{TV} \le nN\cdot c_w^t$, where $\pi_{ij}$ is the law in (i) with $m = m_j$, $g = g_j$ and $p = p_{ij}$.*

*Proof.* Row x of Q is $c_w e_x^T + \mu^T$ by Proposition 1. Since $\mu^T\mathbb{1} = 1 - c_w$, $\Pi = \mathbb{1}\mu^T/(1 - c_w)$ is idempotent and $Q = c_w(I - \Pi) + \Pi$, where $(I - \Pi)$ and Π are complementary idempotents. Hence $Q^t = c_w^t(I - \Pi) + \Pi$, which gives (ii) and (iii), and $\pi^T Q = \pi^T$. For (iii), $e_x^T Q^t - \pi^T = c_w^t(e_x - \pi)^T$, whose TV norm is $c_w^t(1 - \pi(x))$. For (iv), the population chain is a product of nN independent coordinate chains, each with its own law $\pi_{ij}$, and the TV distance between product measures is at most the sum of the marginal distances [19]. ■

**Remark 1.** The decay rate $c_w$ does not depend on m, $C_g$ or $C_p$ separately, and it is far faster than the Doeblin rate $(1 - \alpha)^t$ of Lemma 1. Only the inertia weight $c_w = C_w - C_p$ controls how quickly a solution forgets its current position. When two or three of $g_j$, $p_{ij}$ and $x_{ij}$ coincide, branches of Proposition 1 merge without changing this rate; Section VIII-A works out the effective branch probabilities, and Section IX-G measures how often each case occurs.

## IV. The Elitist Chain

**Theorem 2 (state classification).** *Under A1–A3, a state s = (X, P) is recurrent if and only if every $p_i \in D^*$. The closed classes are $D^N \times \{P\}$ for $P \in (D^*)^N$; all other states are transient. If $D^* = \{x^*\}$, the unique recurrent class is $D^N \times \{(x^*, ..., x^*)\}$ and its stationary distribution is the product of coordinate laws with*

$$\pi_j(x^*_j) = (C_p + c_r/m_j)/(C_p + c_r), \;\; \pi_j(v) = (c_r/m_j)/(C_p + c_r) \text{ for } v \ne x^*_j. \quad (5)$$

*Proof.* If some $p_i \notin D^*$, then by Lemma 1 the next $x'_i$ lies in $D^*$ with positive probability, and then $f(p_i)$ strictly decreases. Since $f(p_i)$ never increases (A3), the chain cannot return to s, so s is transient. If all $p_i \in D^*$, no strict improvement is possible, so P is frozen and X follows the frozen-memory chain, which is irreducible and aperiodic on $D^N$ by Lemma 1; hence $D^N \times \{P\}$ is a closed irreducible class. The stationary law follows from Theorem 1 with $p = g = x^*$: $\mu(x^*_j) = c_g + c_p + c_r/m_j$ and $1 - c_w = C_p + c_r$. ■

**Remark 2.** Once every personal best is optimal and the positions have reached stationarity, coordinate j of each solution equals the optimal value with probability $(C_p + c_r/m_j)/(C_p + c_r)$, which gives an observable diagnostic of the thresholds. At the first hitting time T only g is guaranteed to be optimal, and (5) need not describe the population.

**Theorem 3 (convergence and first-hitting time).** *Let $T = \min\{t : g_t \in D^*\}$ and $q = c_r^n|D^*|/|D| = (1 - C_w)^n|D^*|/|D|$. Under A1–A3, from any initial state*

$$P(T > t) \le (1 - q)^{Nt}, \;\; E[T] \le 1/(1 - (1 - q)^N) < 1 + 1/(Nq). \quad (6)$$

*Proof.* By Lemma 1 and A2, given $s_t$ the N new solutions land in $D^*$ independently, each with probability at least q; if one does, $g_{t+1} \in D^*$ and stays there by A3. Hence $P(T > t + 1 \mid T > t, s_t) \le (1 - q)^N$, and $E[T] = \Sigma_{t\ge 0} P(T > t) \le 1/(1 - (1 - q)^N)$. Finally $(1 - q)^N \le e^{-Nq} < 1/(1 + Nq)$ for $q > 0$. ■

**Proposition 2 (sharpness and a lower bound).** *(i) If $D^* = \{x^*\}$ and $x_i$, $p_i$ and g differ from $x^*$ in every coordinate, then $P(x'_i = x^* \mid s) = q$. (ii) Let $c_w = 0$, let f be a needle function ($D^* = \{x^*\}$ and f is constant on $D \setminus \{x^*\}$), and let every personal best equal a point a that differs from $x^*$ in every coordinate. Then $P(T > t) = (1 - q)^{Nt}$ and $E[T] = 1/(1 - (1 - q)^N)$. (iii) Let $D^* = \{x^*\}$ and $\beta = \prod_j(1 - c_r(m_j - 1)/m_j) \leq (1 - c_r/2)^n$. Then $P(x'_i = x^* \mid s) \leq \beta$ for every state s and every i, and from every initial state with $g_0 \neq x^*$, $P(T > t) \geq (1 - \beta)^{Nt}$ and $E[T] \geq 1/(1 - (1 - \beta)^N) \geq 1/(N\beta)$. Hence the expected number of evaluations, $N(E[T] + 1)$ including the initial population, is at least $1/\beta \geq \exp(nc_r/2)$.*

*Proof.* (i) In Proposition 1 only the uniform term contributes to $y = x^*$. (ii) Before T no strict improvement is possible, so $P = (a, \ldots, a)$ and $g = a$ stay fixed. With $c_w = 0$ the kernel of Proposition 1 does not depend on $x_i$, so each new solution equals $x^*$ with probability exactly $c_r^n/|D| = q$, which is $(c_r/m)^n$ when every $m_j = m$, independently across solutions and generations. Hence T is geometric with success probability $1 - (1 - q)^N$. (iii) In Proposition 1 the copy and inertia branches together put mass at most $1 - c_r$ on any value and the uniform branch puts $c_r/m_j$ on it, so $P(x'_{ij} = x^*_j \mid s) \leq 1 - c_r(m_j - 1)/m_j$; the coordinates are independent given s by A2, so $P(x'_i = x^* \mid s) \leq \beta$. While $g \neq x^*$, no personal best equals $x^*$, so $g_{t+1} = x^*$ only if some new solution equals $x^*$, and by A2 $P(T > t + 1 \mid T > t, s_t) \geq (1 - \beta)^N$. Summing $P(T > t) \geq (1 - \beta)^{Nt}$ over $t \geq 0$ gives the bound on E[T]. Finally $N/(1 - (1 - \beta)^N) \geq 1/\beta$, and $(m_j - 1)/m_j \geq 1/2$ together with $-\ln(1 - z) \geq z$ gives $1/\beta \geq (1 - c_r/2)^{-n} \geq \exp(nc_r/2)$. ■

Proposition 2(i) shows that the one-step bound cannot be improved without information about the state, and (ii) that the geometric bound of Theorem 3 is attained for needle functions without inertia; the simplified bound $1 + 1/(Nq)$ is always strictly larger. On landscapes that carry fitness information the bound is loose (Table IV), because failed generations change the population and raise the later success probabilities. Proposition 2(iii) is a converse that holds on every landscape: with a redraw weight that is held fixed during the run (it may depend on n) and a non-optimal initial global best, the expected cost of reaching a unique optimum grows at least like $\exp(nc_r/2)$. Hence $c_r = O((\log n)/n)$ is necessary, though not sufficient, for a polynomial expected runtime, and the order 1/n of Corollary 1 below cannot be exceeded by more than a log n factor. The bound says nothing about weights that change during a run. For the default $c_r = 0.1$ and $m = 2$ the bound is $1/\beta = 0.95^{-n}$, only about 27 evaluations for $n = 64$ but $1.4 \times 10^{11}$ for $n = 500$. Without inertia it also shows that the exponential factor of the upper bounds (12) and (13) in Section VI cannot be removed: for $c_w = 0$ the redraw share is $r = c_r$, and $1/\beta = \psi^{-n}$ has the same base $\psi = 1 - r(m - 1)/m$ as (12) and (13) (Lemma 2). The next result uses the number of coordinates in which gBest is already correct.

**Theorem 4 (gBest-guided one-step bound).** *Let $D^* = \{x^*\}$ and suppose g differs from $x^*$ in k coordinates. Then for every i,*

$$P(x'_i = x^* \mid s_t) \geq (c_g + c_r/m)^{n-k}(c_r/m)^k. \tag{7}$$

*Proof.* On a coordinate with $g_j = x^*_j$, the probability of $y_j = x^*_j$ is at least $c_g + c_r/m$ by Proposition 1; on the others it is at least $c_r/m$. Multiply using A2. ■

The ratio of this bound to q is $((c_g m + c_r)/c_r)^{n-k}$, which is large once g is close to $x^*$. The bound also exposes a trade-off: $c_r$ repairs wrong coordinates, but every unit of $c_r$ is taken from the mass that preserves correct ones.

**Corollary 1 (end-game redraw weight).** *Take $k = 1$ and $C_g = C_p = C_w = 1 - c_r$. The bound $h(c_r) = (1 - c_r(m - 1)/m)^{n-1}c_r/m$ is maximised at*

$$c_r^* = 1 - C_w^* = m/((m - 1)n), \;\; h(c_r^*) = (1 - 1/n)^{n-1}/((m - 1)n) \geq 1/(e(m - 1)n). \tag{8}$$

*Proof.* With $\beta = (m - 1)/m$, $(\ln h)' = -(n - 1)\beta/(1 - \beta c_r) + 1/c_r$ vanishes at $c_r = 1/(n\beta)$; ln h is concave, so this is the maximum, and $c_r^* \leq 1$ for $n \geq 2$. Substitute, and use $(1 - 1/n)^{n-1} \geq 1/e$. ■

**Remark 3.** For binary variables $c_r^* = 2/n$, so the uniform branch changes a coordinate with probability $c_r^*/2 = 1/n$, the classical mutation rate of the (1+1) EA [16]; precise analyses of optimal fixed and fitness-dependent mutation strengths on OneMax are given in [20], [21]. This is an end-game rule. For general k, the bound $h_k(c_r) = (1 - c_r(m - 1)/m)^{n-k}(c_r/m)^k$ is maximised at $c_r = \min\{1, km/((m - 1)n)\}$, which is $\Theta(1/n)$ only when k is bounded; for $k = \Theta(n)$ the one-step direct-hit maximiser is $\Theta(1)$. Corollary 1 also applies when $C_p > C_g$ late in a run: if $p_i$ and $x_i$ agree with g on its correct coordinates, the mass preserving those coordinates is $c_g + c_p + c_w = 1 - c_r$, and the lower bound $h(c_r)$ of Theorem 4 remains valid; the other references may in addition repair a wrong coordinate.

## V. Binary SSO with Complement Mutation

We also analyse a complement-mutation variant on bits, in which $x'_{ij} = 1 - x_{ij}$ when $\rho \geq C_w$. Uniform redraw changes the bit with probability $c_r/2$; the flip changes it with probability $c_r$. The two variants are not equivalent, and the flip changes the condition for a one-step minorisation.

**Proposition 3 (flip kernel).** *With complement mutation on $S = \{0, 1\}$,*

$$K(y \mid x, p, g) = c_g \cdot 1\{y = g\} + c_p \cdot 1\{y = p\} + c_w \cdot 1\{y = x\} + c_r \cdot 1\{y = 1 - x\} \geq \min(c_w, c_r). \quad (9)$$

*Hence Lemma 1 holds with $c_r/m$ replaced by $\min(c_w, c_r)$, which is positive if and only if $C_p < C_w < 1$. If $C_p = C_w$, no one-step lower bound that is uniform in the state exists.*

*Proof.* Exactly one of $y = x$ and $y = 1 - x$ holds, and the corresponding branch has probability $c_w$ or $c_r$, so $K \geq \min(c_w, c_r)$. If $c_w = 0$ and $g = p = 1 - x$, every branch gives $1 - x$, so $K(x \mid \cdot) = 0$. ■

**Remark 4.** The condition $C_p < C_w < 1$ is sufficient for the one-step minorisation used in Theorem 3, not necessary for convergence. With $C_p = C_w < 1$ the chain may still reach every state over several generations by combining copy and flip steps; only the one-step argument fails. At the other extreme, pure flip ($c_r = 1$) maps every position to its complement, so the positions alternate between two strings.

**Theorem 5 (two-state frozen-memory chain).** *With complement mutation and fixed (p, g), let $\mu = c_g e_g + c_p e_p$. The coordinate chain has stationary law $\pi(v) = (\mu(v) + c_r)/(1 - c_w + c_r)$, eigenvalues 1 and $c_w - c_r$, and*

$$\|e_x^T Q^t - \pi^T\|_{TV} = |c_w - c_r|^t(1 - \pi(x)) = |2C_w - C_p - 1|^t(1 - \pi(x)). \quad (10)$$

*Proof.* $Q = \mathbb{1}\mu^T + c_w I + c_r J$, where J swaps the two states. For $w = (1, -1)^T$, $w^T\mathbb{1} = 0$ and $w^T J = -w^T$, so $w^T Q = (c_w - c_r)w^T$. Balance $\pi(v) = \mu(v) + c_w\pi(v) + c_r\pi(1 - v)$ with $\pi(0) + \pi(1) = 1$ gives π. For a two-state chain the TV distance from π is multiplied by the second eigenvalue in absolute value at each step. ■

**Corollary 2 (hitting time with flip mutation).** *If $c_w > 0$ and $c_r > 0$, Theorem 3 holds with $q = \min(c_w, c_r)^n|D^*| > 0$. If $c_w = 0$ or $c_r = 0$, then $q = 0$ and the argument gives no finite bound (Remark 4). The flip variant has a larger worst-case q than uniform redraw whenever $\min(c_w, c_r) > c_r/2$. If $c_r > c_w$, the second eigenvalue $c_w - c_r$ is negative and each coordinate oscillates around π; if $c_w = c_r$, the frozen-memory chain mixes in one step.*

*Proof.* Replace $c_r/m$ by $\min(c_w, c_r)$ in the proof of Theorem 3, using Proposition 3; the remaining statements follow from Theorem 5. ■

## VI. Expected Runtime on OneMax and LeadingOnes

Theorem 3 bounds the hitting time by that of pure random search. This section uses the fitness-level method [8] to obtain polynomial bounds on two standard test functions. Fix $x^* \in D$ and define, for minimisation,

$$\text{OneMax}_{x^*}(x) = |\{j : x_j \neq x^*_j\}|, \quad \text{LO}_{x^*}(x) = n - \max\{i \leq n : x_j = x^*_j \text{ for all } j \leq i\}. \quad (11)$$

For m = 2 these are the classical OneMax and LeadingOnes functions [16] up to relabelling the bits. T is the first generation in which $f(g) = 0$, and the number of fitness evaluations is $N(T + 1)$.

### A. The gBest owner and a coupling for one block

The analysis rests on one structural fact: g is always the personal best of some solution, its owner. Results for a single particle are likewise transferred to the swarm in the discrete PSO analysis of [11]; here the owner's kernel has the closed form of Theorem 1.

**Lemma 2 (gBest owner).** *Let $i^* = \arg\min_i(f(p_i), i)$, the smallest index among the solutions whose personal best has the best fitness, so that $p_{i^*} = g$ by the tie-breaking rule of Algorithm 1, and let $r = c_r/(1 - c_w) = (1 - C_w)/(1 - C_w + C_p)$ and $\psi = 1 - r(m - 1)/m$. (i) The owner's kernel is $K_j(y \mid x_{i^*}, g, g) = C_p \cdot 1\{y = g_j\} + c_w \cdot 1\{y = x_{i^*,j}\} + c_r/m$; it involves only $x_{i^*}$, g and the owner's own random numbers. With g held fixed it is the chain of Theorem 1 with p = g, whose stationary law puts mass $\psi$ on $g_j$ and r/m on every other value. (ii) Call the pair (g, i*) the global-best record. Whenever the record changes, even if g is unchanged as a vector, the new owner j satisfies $x_j = p_j = g$ in that generation. (iii) If $x_{i^*} = g$, then, given $s_t$, the coordinates of $x'_{i^*}$ are independent, with $P(x'_{i^*,j} = g_j) = 1 - c_r(m - 1)/m \geq \psi$ and $P(x'_{i^*,j} = v) = c_r/m = (1 - c_w)r/m$ for every $v \neq g_j$.*

*Proof.* (i) The gBest and pBest branches of the owner both return $g_j$, so Proposition 1 gives the kernel. Since $\mu(g_j) = C_p + c_r/m$ and $1 - c_w = C_p + c_r$, Theorem 1 gives $\pi(g_j) = (C_p + c_r/m)/(C_p + c_r) = \psi$ and $\pi(v) = r/m$. (ii) The old owner holds the personal best with the smallest fitness and, among ties, the smallest index, and personal bests never get worse. A personal best that did not change in the current generation therefore cannot displace it, so if the record changes, the personal best of the new owner has just changed. A personal best changes only by being replaced by the solution's new position, so $x_j = p_j$. (iii) This is Proposition 1 with x = p = g, and $c_r = (1 - c_w)r \leq r$. ■

**Remark 5.** The quantity r is the share of the redraw branch in the probability mass that does not keep the current value. When $C_p = C_w$ (no inertia), $r = c_r$ and the owner's kernel does not depend on $x_{i^*}$, so in every generation the owner samples g with each coordinate redrawn with probability $c_r$, whatever the other solutions do.

**Lemma 3 (fitness levels with blocks).** *Let $A_0, \ldots, A_L$ partition D so that $f(x) > f(y)$ whenever $x \in A_\ell$, $y \in A_{\ell'}$ and $\ell < \ell'$, with $A_L = D^*$. Suppose that for every $\ell < L$ there are an integer $B_\ell \geq 1$ and $\rho_\ell > 0$ such that, from every state with $g \in A_\ell$, the level of g increases within $B_\ell$ generations with probability at least $\rho_\ell$. Then, from any initial state, $E[T] \leq \Sigma_{\ell<L} B_\ell/\rho_\ell$.*

*Proof.* By A3, $f(g_t)$ never increases, so the level of g never decreases. Split the time that g spends in $A_\ell$ into blocks of $B_\ell$ generations. By the Markov property at the start of each block, every block ends the stay in $A_\ell$ with probability at least $\rho_\ell$, so the number of blocks is stochastically dominated by a geometric variable with mean $1/\rho_\ell$. Summing over the levels gives the claim. With $B_\ell = 1$ this is the usual fitness-level method [8]. ■

**Lemma 4 (coupling for one block).** *Let the levels of Lemma 3 be level sets of f, so that every strict decrease of f(g) increases the level. Fix $\ell < L$ and $B \geq 1$, and suppose that (a) for every $g \in A_\ell$ and every $y_0 \in D$, the chain Y of Lemma 2(i) with p = g and $Y_0 = y_0$ satisfies $P(Y_B \in A_{>\ell}) \geq \rho_a$; and (b) for every $g \in A_\ell$, a solution with x = p = g produces a point of $A_{>\ell}$ in one generation with probability at least $\rho_b$. Then, from every state with $g \in A_\ell$, the level of g increases within B generations with probability at least $\min\{\rho_a, \rho_b\}$.*

*Proof.* Fix a state $s_0$ with global best $g_0 \in A_\ell$ and owner i*, and drive the real process by random numbers $U^{(i)}_t$ that are independent across solutions i and generations t (A2). On the same random numbers define an auxiliary process in which g is held at $g_0$: the owner moves by the chain Y of Lemma 2(i), and every other solution moves and updates its personal best as in Algorithm 1 but with $g_0$ in place of g. Let ν be the first $t \geq 1$ with $f(Y_t) < f(g_0)$, and let σ be the first $t \geq 1$ at which some auxiliary personal best $p_i$, $i \neq i^*$, satisfies $f(p_i) < f(g_0)$, or $f(p_i) = f(g_0)$ with $i < i^*$. The pair $(\nu, Y_B)$ is a function of the owner's random numbers only, and σ of the other solutions' numbers only, so they are independent. Up to $\tau = \min\{\nu, \sigma\}$ the real global-best record (g, i*) equals $(g_0, i^*)$, so the real and auxiliary processes coincide up to and including generation $\min\{\tau, B\}$, and the record changes for the first time at $\tau$; the vector g may stay the same when the record passes to a solution of smaller index. Case $\sigma \ge B$: if $\nu \le B - 1$, the owner improved strictly at $\nu$ and the level increased; if $\nu \ge B$ and $Y_B \in A_{>\ell}$, the real owner samples $Y_B$ at generation B and the level increases. By independence this case contributes at least $P(\sigma \ge B)\cdot P(\{\nu \le B - 1\} \cup \{Y_B \in A_{>\ell}\}) \ge P(\sigma \ge B)\rho_a$. Case $\sigma \le B - 1$: at $\tau \le B - 1$ the record changes, so either f(g) decreases and the level increases, or the record passes to a personal best of equal fitness, possibly equal to $g_0$ itself. In the latter case the new owner satisfies $x = p = g$ at $\tau$ by Lemma 2(ii); this event is determined by the random numbers up to $\tau$, and by the strong Markov property at the stopping time $\tau$ and assumption (b), the level increases at $\tau + 1 \le B$ with conditional probability at least $\rho_b$. This case contributes at least $P(\sigma \le B - 1)\rho_b$. Adding the two cases gives the claim. ■

### *B. Runtime bounds*

Let $B = 1$ if $c_w = 0$ and $B = \max\{1, \lceil \ln n/\ln(1/c_w)\rceil\}$ otherwise, so that $c_w^B \le 1/n$, and let $\theta = \min\{(1 - c_w^B)^n, 1 - c_w\}$. Then $\theta = 1$ when $c_w = 0$, and $\theta \ge \min\{1/4, 1 - c_w\}$ for $n \ge 2$.

**Theorem 6 (OneMax).** *Under A2–A3, for every $N \ge 1$, every $C_g \le C_p \le C_w < 1$ and every initial state, SSO on $OneMax_{x^*}$ satisfies*

$$E[T] \le B\, m\, H_n \,/\, (\theta\, r\, \psi^{n-1}), \quad H_n = \Sigma_{k=1}^{n} 1/k. \tag{12}$$

*Proof.* Use the level sets $A_\ell = \{x : OneMax_{x^*}(x) = n - \ell\}$, and let $k = OneMax_{x^*}(g) \ge 1$. For assumption (a) of Lemma 4, Theorem 1 shows that from any start the B-step law of each coordinate of Y is $\pi + c_w^B(e - \pi) \ge (1 - c_w^B)\pi$, where e is the point mass at $y_0$, and the coordinates are independent. The probability that $Y_B$ equals g except at one of the k coordinates where g is wrong, which is set to $x^*_j$, is therefore at least $\rho_a = k(r/m)\psi^{n-1}(1 - c_w^B)^n$. For assumption (b), Lemma 2(iii) gives the same event in one generation with probability at least $\rho_b = k(r/m)(1 - c_w)\psi^{n-1}$. Hence $\min\{\rho_a, \rho_b\} \ge k\theta r\psi^{n-1}/m$, and Lemmas 3 and 4 with $B_\ell = B$ give $E[T] \le \Sigma_{k=1}^{n} Bm/(k\theta r\psi^{n-1})$. ■

**Theorem 7 (LeadingOnes).** *Under the same conditions, SSO on $LO_{x^*}$ satisfies*

$$E[T] \le (B\, m/(\theta\, r))\, \Sigma_{i=0}^{n-1}\, \psi^{-i}. \tag{13}$$

*Proof.* Use the level sets $A_i = \{x : LO_{x^*}(x) = n - i\}$. If the correct prefix of g has length $i < n$, then $g_{i+1} \ne x^*_{i+1}$, and every point that agrees with g on coordinates 1, …, i and with x* on coordinate i + 1 is strictly better. As in Theorem 6, with this event in place of the OneMax event, $\rho_a \ge (r/m)\psi^i(1 - c_w^B)^{i+1}$ and $\rho_b \ge (r/m)(1 - c_w)\psi^i$, so $\min\{\rho_a, \rho_b\} \ge \theta(r/m)\psi^i$. Apply Lemmas 3 and 4. ■

**Corollary 3 (parameter choice and explicit bounds).** *Let $m \ge 2$ and $n \ge 2$. (i) The bound of Theorem 6 is minimised over r at $r = m/((m - 1)n)$, where $E[T] \le Be(m - 1)nH_n/\theta$. (ii) For $r = am/((m - 1)n)$ with $0 < a \le (m - 1)n/m$, so that $r \le 1$, Theorem 7 gives*

$$E[T_{LO}] \le B(m - 1)n^2(e^{an/(n - a)} - 1)/(a^2\theta), \tag{14}$$

*and at a = 1 also $E[T_{LO}] \le B(m - 1)n((e - 1)n + 1)/\theta$. The leading coefficient $(e^a - 1)/a^2$ is minimised at the root $a^* \approx 1.5936$ of $(a - 2)e^a + 2 = 0$, where it equals 1.544, against $e - 1 \approx 1.718$ at a = 1; this value is admissible whenever $a^* \le (m - 1)n/m$, that is, for $n \ge 4$ when m = 2. (iii) If $C_p = C_w$, then $B = \theta = 1$. (iv) If the inertia weight $c_w \in (0, 1)$ is fixed, then $B = O(\log n)$ and $\theta \ge \min\{1/4, 1 - c_w\}$, so $E[T] = O(n \log^2 n)$ on OneMax and $O(n^2 \log n)$ on LeadingOnes.*

*Proof.* (i) Minimising $1/(r\psi^{n-1})$ over r is the computation in the proof of Corollary 1 with $c_r$ replaced by r. At $r = m/((m - 1)n)$, $m/r = (m - 1)n$ and $\psi^{-(n-1)} = (1 + 1/(n - 1))^{n-1} \le e$. (ii) With $\psi = 1 - a/n$, $\Sigma_{i=0}^{n-1}\psi^{-i} = (\psi^{-n} - 1)(n - a)/a$, and $-\ln(1 - z) \le z/(1 - z)$ gives $\psi^{-n} \le e^{an/(n - a)}$; multiply by $m/r = (m - 1)n/a$ and use $n - a \le n$. At a = 1, $\psi^{-n} = (1 + 1/(n - 1))^n \le en/(n - 1)$, so the sum is at most $(e - 1)n + 1$. Differentiating $(e^a - 1)/a^2$ gives the stated equation. (iii) and (iv) follow from the definitions of B and θ. ■

**Remark 6.** (a) With N = 1, $C_p = C_w$ and m = 2, SSO is the (1+1) EA with strict acceptance and bit-flip probability $p = c_r/2 = a/n$. Its exact expected runtime on LeadingOnes from a uniformly random initial point is $((1 - p)^{1-n} - (1 - p))/(2p^2) \approx n^2(e^a - 1)/(2a^2)$ [17], so the bound of Corollary 3(ii) is asymptotically twice the exact value in this case, and $a^* \approx 1.59$ is the optimal rate found in [17]; on OneMax, $enH_n$ is the classical fitness-level bound [8], [16]. These cases are consistency checks; the general results concern $N > 1$, $C_g \neq C_p$ and $c_w > 0$. (b) The inertia weight enters only through B and θ, that is, through the time the owner needs to forget its position, which Theorem 1 gives exactly. The extra log n factor in Corollary 3(iv) is probably an artefact of the block argument; Section IX-D finds the same scaling for $c_w = 0.1$ as for $c_w = 0$, and Section VII shows that, for a frozen memory and stationary positions, inertia cannot shorten first-entry times relative to independent sampling from the same stationary law, a comparison that need not hold from non-stationary positions (Example 1). (c) The bounds do not improve with N, because only the owner is guaranteed to search close to g; the other solutions copy g only with probability $C_g$ per coordinate. In the experiments the number of generations falls roughly in proportion to 1/N, so the bound $N(T + 1)$ on evaluations is pessimistic for $N > 1$; proving this speed-up is left open. (d) The parameter values in Corollary 3 minimise upper bounds, not the exact runtime of the swarm; Section IX-E compares them with simulations. (e) For complement mutation (Section V) the owner's kernel remains dependent on its current position $x_{i^*}$ even when $C_p = C_w$. Extending the block argument would require separate accessibility and mixing conditions, and such runtime bounds are not established here. Some condition is needed: with pure flip ($c_g = c_p = c_w = 0$) the positions alternate between two complementary strings (Remark 4), so for OneMax with $n = 2$, $N = 1$ and $x_0 = p_0 = g_0 = 01$ the optimum 00 is never produced.

## VII. What the Inertia Weight Does

The inertia branch is the only part of the SSO rule that depends on the current position $x_i$. In the runtime bounds of Section VI it enters only through the time the owner needs to forget its position (Remark 6(b)). This section makes precise what inertia does when the memory is fixed: it leaves the long-run sampling law unchanged after normalisation, a solution started at its personal best is at every time stochastically no farther from it than under the stationary law, and from stationary positions it cannot speed up the first entry into any target set. Example 1 shows that the last statement does not extend to non-stationary positions.

**Theorem 8 (inertia under frozen memory).** *Fix the memory (g, P) and let $c_w < 1$. For one solution i let $x_0 = x_i$ and let $x_1, x_2, \ldots$ be its successive positions while the memory stays fixed, and let $\pi = \otimes_j \pi_j$ be the stationary law of Theorem 1. (i) $\pi_j$ is the law of one update without inertia, with branch probabilities $(c_g, c_p, 0, c_r)/(1 - c_w)$. (ii) For every $t \geq 0$ the coordinates of $x_t$ are independent, and $P(x_{tj} = y) = c_w^t 1\{y = x_{0j}\} + (1 - c_w^t)\pi_j(y)$. In particular, if $x_0 = p_i$, then $P(x_{tj} \neq p_{ij}) = (1 - c_w^t)\pi_j(S_j \setminus \{p_{ij}\})$, which increases with t towards its stationary value; for the owner ($g = p_i$) it equals $(1 - c_w^t)r(m_j - 1)/m_j$. (iii) If $x_0 \sim \pi$, then for every $A \subseteq D$ and every $t \geq 1$*

$$P(x_0 \notin A, \ldots, x_{t-1} \notin A) \geq (1 - \pi(A))^t, \tag{15}$$

*so the number of updates $N_A = \min\{t \geq 1 : x_t \in A\}$ until a position in A is produced, for which $P(N_A > t) = P(x_1, \ldots, x_t \notin A)$ equals the left-hand side by stationarity, stochastically dominates the geometric law with success probability $\pi(A)$, which is its exact law without inertia, and its mean is at least $1/\pi(A)$. Given the memory, the N solutions move independently; if their initial positions are independent with laws $\pi^{(1)}, \ldots, \pi^{(N)}$, the same holds for the first solution of the swarm that enters A, with $(1 - \pi(A))^t$ replaced by $\prod_i(1 - \pi^{(i)}(A))^t$.*

*Proof.* (i) An update with $c_w = 0$ and branch probabilities divided by $1 - c_w$ gives y the probability $(c_g 1\{y = g_j\} + c_p 1\{y = p_{ij}\} + c_r/m_j)/(1 - c_w) = \pi_j(y)$ of Theorem 1, whatever $x_{ij}$ is. (ii) $\Pi_j$ is idempotent and $\Pi\Pi_j = \Pi_j$, so $Q_j^t = c_w^t I + (1 - c_w^t)\Pi_j$, and the coordinates are updated independently (A2). (iii) Each $Q_j$ satisfies $\pi_j(x)Q_j(x, y) = c_w\pi_j(x)1\{x = y\} + (1 - c_w)\pi_j(x)\pi_j(y)$, which is symmetric in x and y, so $Q = \otimes_j Q_j$ is reversible with respect to π; as an operator on $L^2(\pi)$ it is self-adjoint, with eigenvalues that are products of 1 and $c_w$, hence in [0, 1]. Let $B = D \setminus A$ and let $Q_B$ be the restriction of Q to B. It is self-adjoint on $L^2(\pi|_B)$ and positive semidefinite, being a compression of a positive semidefinite operator, so $f(t) := P(x_0, \ldots, x_{t-1} \in B) = \langle 1_B, Q_B^{t-1} 1_B \rangle_\pi = \Sigma_k a_k \lambda_k^{t-1}$ with $a_k \geq 0$, $\Sigma_k a_k = \pi(B)$ and $\lambda_k \in [0, 1]$ [19]. Jensen's inequality for the convex map λ $\mapsto \lambda^{t-1}$ gives $f(t) \geq \pi(B)(f(2)/\pi(B))^{t-1}$. Finally $f(2) - \pi(B)^2 = \langle h, Qh\rangle_\pi \geq 0$ for $h = 1_B - \pi(B)$, so $f(2) \geq \pi(B)^2$ and $f(t) \geq \pi(B)^t$. Summing $P(N_A > t) = f(t)$ over $t \geq 0$ gives the mean. Given the memory and independent initial positions, the solutions evolve independently, which gives the swarm statement. ■

**Example 1 (stationarity cannot be dropped).** *Let $m = n = 2$, $N = 1$, $g = p_1 = 10$, and $f(00) = f(11) = 0$, $f(10) = 1$, $f(01) = 2$, so that $A = \{00, 11\}$ is the set of improving points and the memory stays fixed until A is entered. The settings $(c_g, c_p, c_w, c_r) = (0.8, 0, 0, 0.2)$ and $(0.56, 0, 0.3, 0.14)$ have the same normalised probabilities, hence the same stationary law $\pi = (0.09, 0.01, 0.81, 0.09)$ on $(00, 01, 10, 11)$, and $\pi(A) = 0.18$. From $x_0 = 01$ the expected number of updates until A is entered is $1/\pi(A) = 50/9 \approx 5.56$ without inertia and $3800/819 \approx 4.64$ with inertia.*

*Proof.* Without inertia the positions $x_1, x_2, \ldots$ are independent with law $\pi$. With inertia each coordinate keeps its value with probability $0.3 + 0.7\pi_j(x_{0j})$; the restriction of the kernel to $B = \{01, 10\}$ has rows (0.1369, 0.3969) and (0.0049, 0.8649), and solving $(I - Q_B)h = 1$ gives $h(01) = 3800/819$. Both values were also checked in exact rational arithmetic. ■

From $x_0 = 01$, inertia keeps one coordinate while the other moves, which produces the points of A directly; without inertia each position is drawn afresh and is most likely 10. Theorem 8(iii) is therefore a statement about stationary positions, and the effect of inertia from the positions that actually occur in a run has to be measured, as Sections IX-F and IX-H do.

**Remark 7.** Theorem 8 and Example 1 separate the effects of the inertia weight. By (i), inertia does not change where a solution samples once the memory has been fixed for a while: the stationary law depends only on the normalised probabilities, and for the owner only on the redraw share r. By (ii), inertia does not move a solution farther from its memory than the stationary law either: starting from the personal best, the probability that a coordinate has moved never exceeds its stationary value, so, the coordinates being independent for a fixed memory, the distance to the personal best is at every time stochastically dominated by its stationary distribution. What inertia adds is positive correlation between successive positions; from stationary positions (iii) shows that this correlation cannot speed up the first entry into any set, whatever its shape. From non-stationary positions it can (Example 1), and in a run the positions are rarely stationary: after an improvement $x_i = p_i$, and after a change of g the positions still reflect the old memory. In a run the effect of inertia therefore combines three components: the correlation at stationarity, which cannot speed up the search for a fixed memory; the transient effect of the actual positions, which can go either way; and the influence of inertia on which memories the swarm builds. Sections IX-F to IX-H separate them.

## VIII. How the Three Reference Values Combine

Proposition 1 builds each new coordinate from three reference values, $g_j$, $p_{ij}$ and $x_{ij}$, and a uniform redraw. This section studies the two extreme situations: the reference values agree, and branches merge (Section VIII-A); or they disagree, and the copy branches recombine information held by different solutions (Section VIII-B).

### *A. Coincident reference values*

The four branches of (1) return different values only when $g_j$, $p_{ij}$ and $x_{ij}$ are distinct. When two or three of them coincide, the branches that return the same value act as one, and the effective values of $c_g$, $c_p$ and $c_w$ change; $c_r$ never does.

**Proposition 4 (effective branch probabilities).** *Fix i and j and let $m = m_j$. The kernel $K_j(\cdot \mid x_i, p_i, g)$ of Proposition 1 equals the same kernel with $(c_g, c_p, c_w, c_r)$ replaced by the effective values listed below. (A) $g_j$, $p_{ij}$ and $x_{ij}$ distinct: $(c_g, c_p, c_w, c_r)$. (B) $g_j = p_{ij} \neq x_{ij}$: $(C_p, 0, c_w, c_r)$. (C) $p_{ij} = x_{ij} \neq g_j$: $(c_g, 0, c_p + c_w, c_r)$. (D) $g_j = x_{ij} \neq p_{ij}$: $(0, c_p, c_g + c_w, c_r)$. (E) $g_j = p_{ij} = x_{ij}$: $(0, 0, C_w, c_r)$. In every case*

$$P(x'_{ij} \neq x_{ij} \mid s_t) = c_g \cdot 1\{g_j \neq x_{ij}\} + c_p \cdot 1\{p_{ij} \neq x_{ij}\} + c_r(m-1)/m, \tag{16}$$

*and, with P and g fixed, the coordinate chain keeps the eigenvalue $c_w$ of Theorem 1: coincidences merge atoms of the stationary law but do not change the decay rate.*

*Proof.* In Proposition 1, indicator terms with the same argument add; in case (C), for example, $c_p \cdot 1\{y = p_{ij}\} + c_w \cdot 1\{y = x_{ij}\} = (c_p + c_w) \cdot 1\{y = x_{ij}\}$. The change probability is $1 - K_j(x_{ij} \mid \cdot) = 1 - c_g \cdot 1\{g_j = x_{ij}\} - c_p \cdot 1\{p_{ij} = x_{ij}\} - c_w - c_r/m$, which equals the expression shown because $c_g + c_p + c_w + c_r = 1$. The eigenvalues in Theorem 1 do not depend on g or p, so they are the same when g = p. ■

The effective values are a state-dependent representation of the current transition row of one coordinate, not a permanent replacement of the parameters of the algorithm; the frozen-memory spectrum remains that of the original kernel. For example, with $(c_g, c_p, c_w, c_r) = (0.5, 0.3, 0.1, 0.1)$ and $g_j = p_{ij} = x_{ij}$, the current row equals that of (0, 0, 0.9, 0.1), but the coordinate chain still decays at rate $c_w = 0.1$, whereas a chain run permanently with (0, 0, 0.9, 0.1) would decay at rate 0.9 once $x_{ij}$ leaves $g_j$.

The three coincidences of whole vectors that occur in every run, and a fourth that needs a tie, read as follows; the numbers are for the default setting $(c_g, c_p, c_w, c_r) = (0.5, 0.3, 0.1, 0.1)$ used in Section X.

- **$g = p_i$: solution i holds a globally best personal best.** It is the owner i* of Lemma 2 unless a solution of smaller index holds the same vector. Every coordinate is in class B or E, so the owner runs SSO with $(C_p, 0, c_w, c_r)$, here (0.8, 0, 0.1, 0.1): $c_g$ absorbs $c_p$, while $c_w$ and $c_r$, and hence the redraw share $r = c_r/(1 - c_w)$, are unchanged. For the owner this is Lemma 2(i), and it is why $c_g$ and $c_p$ enter the runtime bounds of Section VI only through $C_p$.
- **$p_i = x_i$, for example just after an improvement.** Algorithm 1 sets $p_i \leftarrow x'_i$ and $X \leftarrow X'$, so every solution whose personal best improved, and every solution at t = 0, starts the next generation in this state; the state also persists after an update that leaves $x_i$ unchanged. Its coordinates are in class C or E, and it runs SSO with $(c_g, 0, c_p + c_w, c_r)$, here (0.5, 0, 0.4, 0.1): $c_w$ absorbs $c_p$, while $c_g$ and $c_r$ are unchanged. A larger $c_p$ therefore acts as extra inertia for a solution in this state and leaves only $c_g + c_r/m$ to move a coordinate towards $g_j$. The extra inertia belongs to the state, not to the chain: once $x_{ij}$ differs from $p_{ij}$, the pBest branch acts on its own again, and the decay rate of Theorem 1 remains $c_w$.
- **$g = p_i = x_i$, for example just after the owner improved.** Every coordinate is in class E, with effective values $(0, 0, C_w, c_r)$, here (0, 0, 0.9, 0.1): $c_g$, $c_p$ and $c_w$ act only through their sum $C_w = 1 - c_r$. The new solution is g with each coordinate redrawn independently with probability $c_r$, a uniform mutation that changes each coordinate with probability $c_r(m - 1)/m$. By Lemma 2(ii), every change of g creates this state, although the state can also persist without a change of g. It is the setting of Corollary 1, whose end-game value $c_r = m/((m - 1)n)$ brings the expected number of changed coordinates to about one. Because the redraw branch is then the only source of change, $c_r$ also limits jumps: a given point at distance k from g is produced with probability $(c_r/m)^k(1 - c_r(m - 1)/m)^{n-k}$, which is maximised at $c_r = \min\{1, km/((m - 1)n)\}$ (Remark 3). The end-game value therefore trades the ability to change several coordinates at once for fast single-coordinate repair; Section X-B shows the cost on $\text{Jump}_2$. Coincidence of the reference values thus identifies the redraw branch as the only source of change; it does not by itself justify reducing its probability.
- **$g = x_i \neq p_i$.** For whole vectors this needs $f(p_i) = f(g)$ with $p_i \neq g$, because otherwise $p_i$ would have been replaced by $x_i = g$; for single coordinates, class D is common. The effective values are $(0, c_p, c_g + c_w, c_r)$, and the pBest branch is the only copy branch that moves the coordinate away from $g_j$, towards $p_{ij}$; the redraw branch can also move it away.

**Binary variables.** For m = 2, at least two of $g_j$, $p_{ij}$ and $x_{ij}$ coincide on every coordinate, so class A cannot occur and the four branches never return four different values. On bits the pBest branch acts as a gBest copy (class B), as inertia (class C) or as a pull away from $g_j$ (class D), and has no separate role in class E.

**What this means for each parameter.** (i) $c_r$ is the only branch probability that is the same in all five classes. It is the only source of change in class E and therefore governs the search once the references agree on most coordinates, which Section IX-G shows to be the typical situation; this is why the runtime bounds and the end-game rule are expressed through it. (ii) $c_w$ acts as inertia of its own only in classes A and B, that is, on coordinates where $x_{ij}$ has drifted away from the reference values; elsewhere it merges with $c_p$ (class C), with $c_g$ (class D) or with both (class E). This matches Remark 6(b), and Theorem 8 shows that, for a fixed memory and stationary positions, this inertia cannot speed up the search. (iii) $c_g$ is the only copy branch that pulls towards $g_j$ when $p_{ij} = x_{ij}$ (class C), and for the owner it is merged with $c_p$ into $C_p$ (class B). (iv) $c_p$ acts on its own only in classes A and D. Moving probability from $c_g$ to $c_p$ therefore leaves the owner unchanged but slows the spread of g to solutions with $p_i = x_i$, where $c_p$ acts as inertia.

### *B. Disagreeing reference values: implicit recombination*

Where the reference values differ, the copy branches combine them. Proposition 1 gives the probability of every offspring in closed form, and evaluating it at a target point shows which coordinates have to be produced by the redraw branch.

**Proposition 5 (recombination bound).** *For every state s, every i and every y ∈ D,*

$$P(x'_i = y \mid s) = \prod_j (c_g \cdot 1\{g_j = y_j\} + c_p \cdot 1\{p_{ij} = y_j\} + c_w \cdot 1\{x_{ij} = y_j\} + c_r/m_j). \tag{17}$$

*If $c_r = 1$, every offspring is uniform on D and $P(x'_i = y \mid s) = 1/|D|$. Otherwise, for each j let $a_j = c_g \cdot 1\{g_j = y_j\} + c_p \cdot 1\{p_{ij} = y_j\} + c_w \cdot 1\{x_{ij} = y_j\}$, the probability that an active copy branch supplies $y_j$. Let I be the number of coordinates with $a_j = 0$, J the number with $0 < a_j < 1 - c_r$, $m = \max_j m_j$, and γ the smallest positive value among $c_g$, $c_p$ and $c_w$, plus $c_r/m$. Then $P(x'_i = y \mid s) \geq (1 - c_r(m - 1)/m)^{n-I-J} \gamma^J (c_r/m)^I$, and the I coordinates can only be produced by the redraw branch.*

*Proof.* The product form is Proposition 1 with A2, and the factor of coordinate j is $a_j + c_r/m_j$. If $a_j = 1 - c_r$, it equals $1 - c_r + c_r/m_j \geq 1 - c_r(m - 1)/m$; if $a_j = 0$, it equals $c_r/m_j \geq c_r/m$, and only a redraw can produce $y_j$; otherwise $a_j$ is at least the smallest positive branch probability, so the factor is at least γ. ■

Theorem 4 is the special case in which only g is used: it pays $c_r/m$ on every coordinate where g is wrong. Proposition 5 pays this price only where no active branch holds the target value, so information held by $p_i$, or by $x_i$ when $c_w > 0$, can be used without a redraw. The effect is largest on functions with a fitness gap.

**Corollary 4 (complementary local optima of Jump functions).** *Let m = 2, let g and $p_i$ be at Hamming distance k from x* with disjoint sets of wrong coordinates, and let $x_i = p_i$. Then*

$$P(x'_i = x^* \mid s) = (c_p + c_w + c_r/2)^k (c_g + c_r/2)^k (1 - c_r/2)^{n-2k}, \tag{18}$$

*whereas a solution whose three references all equal one point at distance k from x* produces x* with probability $(c_r/2)^k(1 - c_r/2)^{n-k}$.*

*Proof.* On the k coordinates where g is wrong, $p_{ij} = x_{ij} = x^*_j$; on the k coordinates where $p_i$ is wrong, only $g_j = x^*_j$; on the other n − 2k coordinates all three references are correct. Apply the product in Proposition 5. In the second case the k wrong coordinates need the redraw branch and the others must not be redrawn to the wrong value. ■

**Remark 8.** On $\text{Jump}_k$ the local optima are the points at distance k from x*, and the points in between have lower fitness, so an elitist solution at a local optimum must produce x* in one step. For fixed k, $c_r = \Theta(1/n)$ and copy weights satisfying $c_g \geq \delta$ and $c_p + c_w \geq \delta$ for a constant δ > 0, the first probability in Corollary 4 is Ω(1), whereas the second is $\Theta(n^{-k})$; without such weights the advantage can vanish, and with $c_g = 0$ the two probabilities coincide. Under these conditions two complementary memories make the gap as easy to cross as a constant number of correct copies, a form of uniform crossover between g and $p_i$, related to the crossover speed-ups proved for the (μ+1) GA [14], [15]. For $(c_g, c_p, c_w, c_r) = (0.5, 0.3, 0.1, 0.1)$, k = 3 and n = 20 the two probabilities are $7.39\times10^{-3}$ and $5.23\times10^{-5}$. Corollary 4 does not say how often complementary memories arise, which depends on the swarm dynamics; by Remark 7 this is also one of the routes by which inertia could help. Sections IX-G and IX-H measure both.

## IX. Numerical Validation

All computations use the code and per-run data provided as supplementary material; the simulations use fixed random seeds. The output of the code agrees with exact values wherever these are available, namely the exact references in Tables VIII and XII.

### *A. Closed forms of Theorems 1 and 5*

Table III compares the closed forms with direct linear algebra on the coordinate matrices, including $m > 2$, $p \neq g$ and the three regimes of the flip variant. The second eigenvalue agrees with $c_w$ (Theorem 1) or $c_w - c_r$ (Theorem 5), and the stationary laws and TV distances at $t \in \{1, 2, 5, 10\}$ agree to machine precision.

TABLE III. Closed forms versus direct computation

| Kernel | Setting | Second eigenvalue | Predicted | Max error in π | Max error in TV |
|---|---|---|---|---|---|
| Uniform redraw (Thm. 1) | $m = 3$, $(C_g, C_p, C_w) = (0.3, 0.5, 0.8)$, $p \neq g$ | 0.3000 | 0.3000 | 1e-17 | 8e-17 |
| Uniform redraw (Thm. 1) | $m = 5$, $(C_g, C_p, C_w) = (0.2, 0.6, 0.9)$, $p \neq g$ | 0.3000 | 0.3000 | 6e-17 | 1e-16 |
| Uniform redraw (Thm. 1) | $m = 5$, $(C_g, C_p, C_w) = (0.4, 0.4, 0.7)$, $p = g$ | 0.3000 | 0.3000 | 1e-16 | 8e-17 |
| Uniform redraw (Thm. 1) | $m = 8$, $(C_g, C_p, C_w) = (0.1, 0.3, 0.95)$, $p \neq g$ | 0.6500 | 0.6500 | 1e-16 | 6e-16 |
| Flip (Thm. 5) | $(c_w, c_r) = (0.1, 0.4)$, $c_w<c_r$ | -0.3000 | -0.3000 | 6e-17 | 4e-17 |
| Flip (Thm. 5) | $(c_w, c_r) = (0.25, 0.25)$, $c_w=c_r$ | -0.0000 | 0.0000 | 6e-17 | 1e-16 |
| Flip (Thm. 5) | $(c_w, c_r) = (0.5, 0.2)$, $c_w>c_r$ | 0.3000 | 0.3000 | 0e+0 | 7e-17 |

### *B. Exact first-hitting time on a small binary instance*

For $n = 3$, $m = 2$ and $N = 2$ the state space has 4096 states. We used $f = (3.0, 2.0, 2.5, 1.0, 1.5, 0.5, 2.2, 0.0)$ on 000, …, 111 in lexicographic order, so $x^* = 111$ and the landscape is partly deceptive. The target set is $\{s : g(s) \in D^*\}$; the expected hitting times h solve $(I - Q_{nonhit})h = \mathbb{1}$, where $Q_{nonhit}$ is the transition matrix restricted to the 3136 states with $g \notin D^*$. (The elitist chain has 4032 transient states in all, but T stops as soon as g is optimal.) "Uniform start" means $x_{1,0}$, $x_{2,0}$ drawn independently and uniformly from D with $P_0 = X_0$; "worst state" is the maximum of h over all non-hit states (Table IV, Fig. 1).

TABLE IV. Exact expected first-hitting time (generations) versus the bound of Theorem 3

| $(C_g, C_p, C_w)$ | q | Exact E[T], uniform start | Exact E[T], worst state | Bound $1/(1 - (1 - q)^N)$ |
|---|---|---|---|---|
| (0.4, 0.6, 0.8) | $1.00\times10^{-3}$ | 6.29 | 12.48 | 500.3 |
| (0.3, 0.5, 0.7) | $3.38\times10^{-3}$ | 4.78 | 9.08 | 148.4 |
| (0.4, 0.6, 0.9) | $1.25\times10^{-4}$ | 11.10 | 23.05 | 4000.3 |
| (0.2, 0.4, 0.5) | $1.56\times10^{-2}$ | 3.61 | 6.26 | 32.3 |
| (0.45, 0.7, 0.95) | $1.56\times10^{-5}$ | 20.63 | 43.80 | 32000.3 |

The bound holds in every case and is loose by one to three orders of magnitude on this landscape. For the needle function with $c_w = 0$ and both personal bests at 000, the exact E[T] equals $1/(1 - (1 - q)^N)$ to within $10^{-11}$ from every $X_0$, for $(C_g, C_p, C_w) = (0.3, 0.6, 0.6), (0, 0.5, 0.5), (0.5, 0.8, 0.8)$ (E[T] = 62.75, 32.25, 500.25), as Proposition 2(ii) states; in these settings it is also the maximum over all non-hit states.

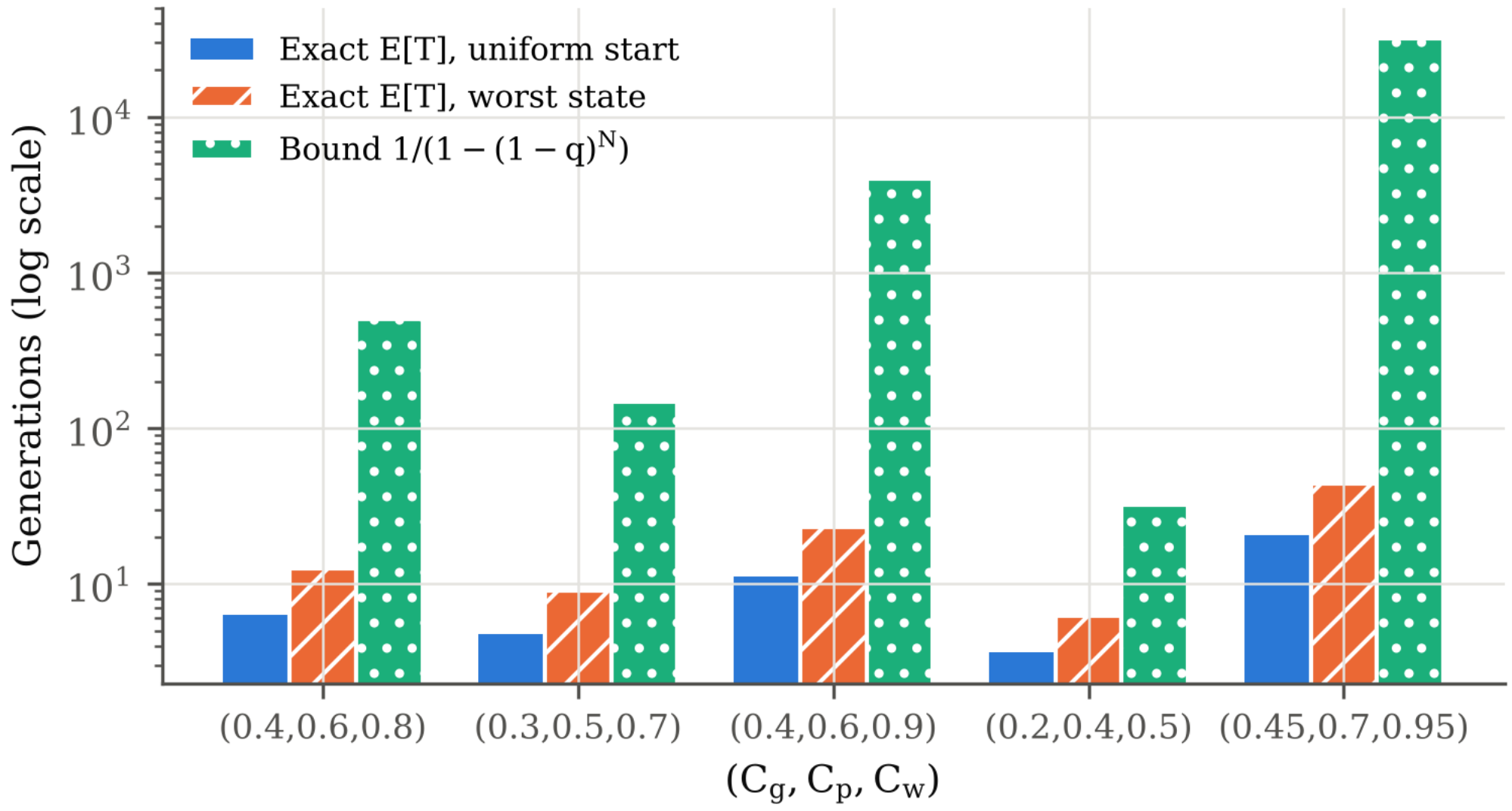

Fig. 1. Exact expected hitting time from a uniform start and from the worst state, compared with the bound of Theorem 3 (log scale).

### *C. The end-game rule on OneMax*

This experiment checks the end-game rule of Corollary 1 over full runs; it is not a direct test of Corollary 1, which concerns one generation at distance one. Binary SSO (uniform redraw) was run on OneMax with $N = 10$, $C_g = 0.6C_w$, $C_p = 0.9C_w$, 200 runs per setting and $c_r = \kappa/n$. Here $c_w = 0.1(1 - c_r)$, so $r = c_r/(0.9 + 0.1c_r)$, and the value $r = 2/n$ of Corollary 3(i) corresponds to $\kappa = 1.8n/(n - 0.2) \approx 1.81$. The lowest sample mean on the tested grid lies at $\kappa \in [1.5, 2.5]$ for $n = 20$, 40 and 80, and the means for $\kappa \in [1.5, 3]$ are within 9% of that minimum (Fig. 2, Table V).

TABLE V. OneMax: $\kappa = n(1 - C_w)$ with the lowest sample mean on the tested grid (mean evaluations ± standard error, 200 runs)

| n | κ with lowest mean | Mean at that κ | Mean at κ = 2 | Ratio |
|---|---|---|---|---|
| 20 | 1.5 | 143.2 ± 3.9 | 146.5 ± 4.1 | 1.023 |
| 40 | 2 | 355.4 ± 7.9 | 355.4 ± 7.9 | 1.000 |
| 80 | 2.5 | 848.5 ± 17.9 | 863.3 ± 18.1 | 1.017 |

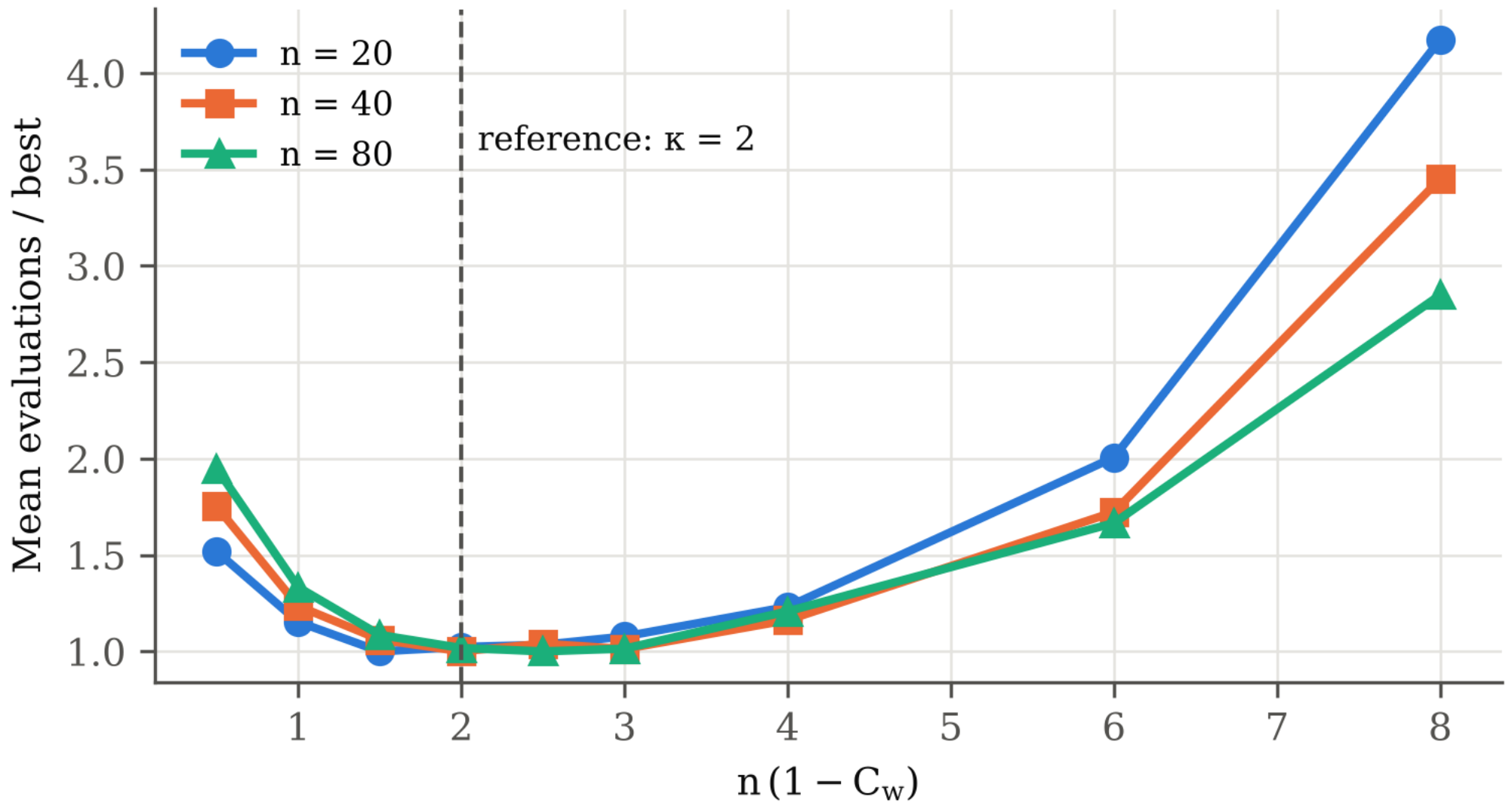

Fig. 2. OneMax: mean evaluations to the optimum, normalised by the lowest mean for each n, against $n(1 - C_w)$. The dashed line marks $\kappa = 2$.

### *D. Expected runtime on OneMax and LeadingOnes*

Binary SSO with uniform redraw was run with $N = 10$ and $r = 2/n$, in two settings: no inertia ($C_p = C_w$, so $c_r = 2/n$) and $c_w = 0.1$ (so $c_r = 1.8/n$). In both, $C_g = C_p/2$. Each entry is the mean over 100 runs from a uniformly random population with $P_0 = X_0$. Table VI and Fig. 3 compare the mean number of generations with the bounds of Corollary 3 (at a = 1 for LeadingOnes).

TABLE VI. Mean generations to the optimum (± standard error, 100 runs, N = 10) versus the bounds of Corollary 3

| Problem | n | Observed, $c_w = 0$ | Bound, $c_w = 0$ | Observed, $c_w = 0.1$ | Bound, $c_w = 0.1$ |
|---|---|---|---|---|---|
| OneMax | 16 | 9.5 ± 0.3 | 147 | 10.7 ± 0.4 | 345 |
| OneMax | 32 | 24.9 ± 0.7 | 353 | 24.5 ± 0.7 | 974 |
| OneMax | 64 | 58.2 ± 1.8 | 825 | 65.8 ± 2.2 | 3140 |
| OneMax | 128 | 140.1 ± 3.9 | 1890 | 156.2 ± 4.8 | 6446 |
| OneMax | 256 | 340.8 ± 9.1 | 4262 | 351.5 ± 10.5 | 16518 |
| LeadingOnes | 16 | 20.7 ± 0.9 | 456 | 24.1 ± 1.1 | 1071 |
| LeadingOnes | 32 | 81.8 ± 2.4 | 1792 | 87.3 ± 2.4 | 4942 |
| LeadingOnes | 64 | 294.6 ± 6.1 | 7102 | 344.4 ± 8.0 | 27025 |
| LeadingOnes | 128 | 1302.4 ± 22.2 | 28280 | 1390.6 ± 25.8 | 96432 |

Every observed mean lies below its bound, by a factor between 13 and 78. Log–log regressions of the mean on n give slopes 1.28 ($c_w = 0$) and 1.27 ($c_w = 0.1$) for OneMax, close to the local slope $1 + 1/\ln n \approx 1.2$–$1.4$ of $n \ln n$ over this range, and 1.98 and 1.95 for LeadingOnes. The ratios $T/(n \ln n)$ and $T/n^2$ of the unrounded means lie in [0.215, 0.252] and [0.071, 0.094], with the endpoints rounded outward to three decimals. With $c_w = 0.1$ the runtime is at most 17% above that with $c_w = 0$, which supports Remark 6(b): the extra log n factor of Corollary 3(iv) is not visible in these experiments.

With a single solution the bound is close: for $C_p = C_w$, $n = 64$ and $N = 1$ the mean was 587 generations against the bound 825. For N = 5, 10 and 20 the mean fell to 114, 58 and 34 generations, roughly in proportion to 1/N, as noted in Remark 6(c).

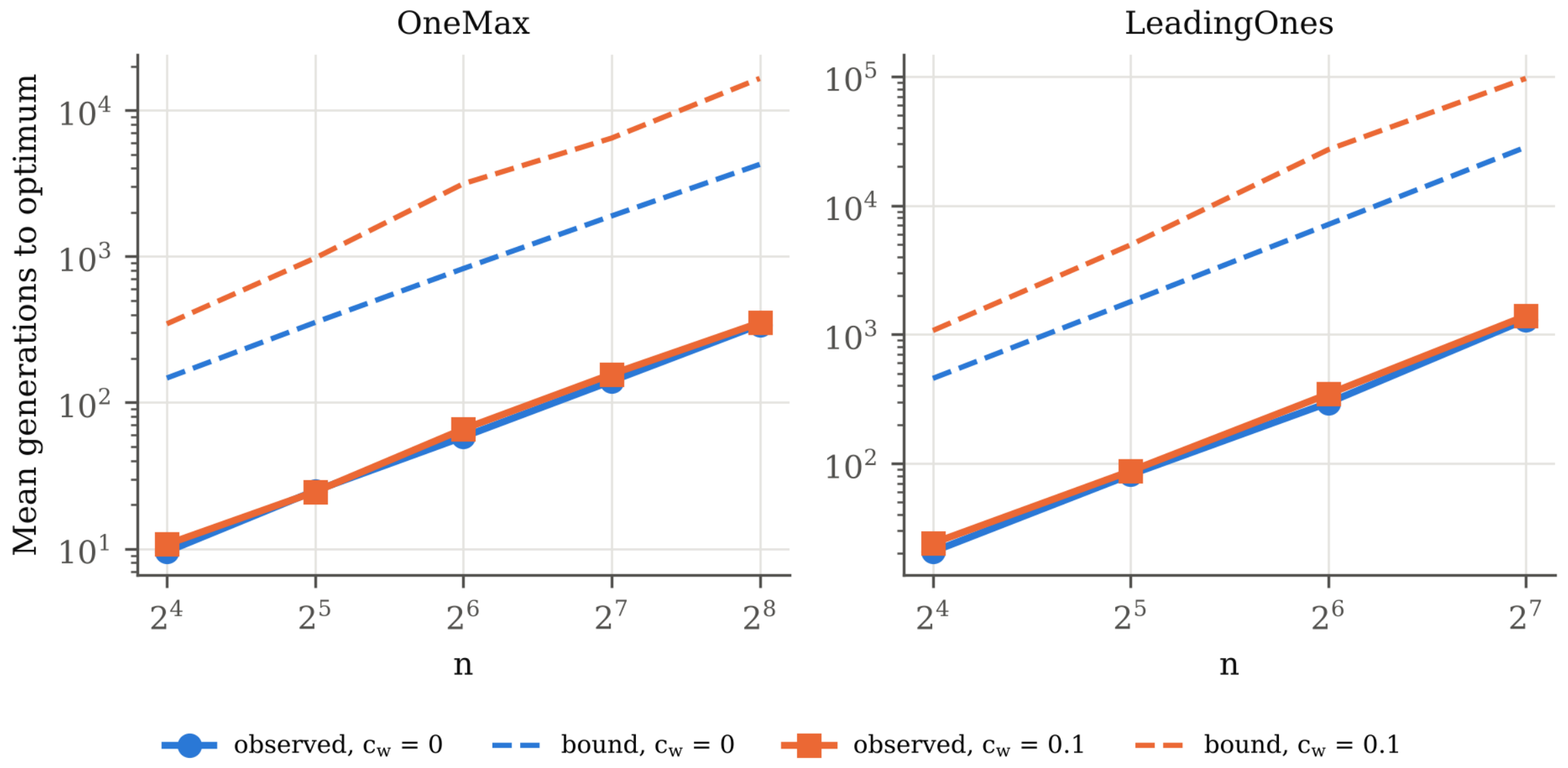

Fig. 3. Mean generations to the optimum on OneMax (left) and LeadingOnes (right), N = 10, for $c_w = 0$ and $c_w = 0.1$ (markers), with the bounds of Corollary 3 (dashed lines); both axes are logarithmic.

### *E. Parameter values: bounds versus simulation*

To compare the parameter values of Corollary 3 with full runtimes, binary SSO without inertia ($C_p = C_w$, $C_g = C_p/2$) was run on OneMax and LeadingOnes with n = 64, bit-change probability a/n (r = 2a/n), N ∈ {1, 10} and 100 runs per setting (Table VII, Fig. 4). For N = 1 the LeadingOnes means agree with the exact (1+1) EA runtime of [17] within two standard errors at every a, and the lowest means lie at a = 1.5–1.75, near a* ≈ 1.59. For OneMax with N = 1 the lowest mean is at a = 1.5, but the pairwise differences of the means for a ∈ [1, 1.75] are below two standard errors, consistent with the bound minimiser a = 1. For N = 10 the lowest mean moves to a = 1.25 on OneMax, while on LeadingOnes it lies at a = 1.5, with all means for a ∈ [1, 1.75] within 3.3% of the lowest. The values of Corollary 3 are therefore good defaults, but they minimise upper bounds and not the exact runtime of the swarm.

TABLE VII. Mean generations (± standard error, 100 runs, n = 64, $C_p = C_w$) against the bit-change parameter a

| a | OneMax, N = 1 | OneMax, N = 10 | LeadingOnes, N = 1 | (1+1) EA exact [17] | LeadingOnes, N = 10 |
|---|---|---|---|---|---|
| 0.5 | 775.8 ± 23.2 | 73.8 ± 2.5 | 5312.8 ± 113.9 | 5299.1 | 458.1 ± 12.0 |
| 0.75 | 697.7 ± 23.2 | 62.0 ± 1.8 | 4152.9 ± 89.3 | 4053.0 | 343.7 ± 8.0 |
| 1 | 628.8 ± 22.4 | 63.9 ± 2.1 | 3542.6 ± 68.4 | 3507.5 | 314.7 ± 8.2 |
| 1.25 | 649.4 ± 26.9 | 54.3 ± 1.6 | 3310.7 ± 72.1 | 3256.2 | 310.6 ± 6.7 |
| 1.5 | 590.7 ± 24.0 | 55.8 ± 1.9 | 3109.2 ± 72.0 | 3166.6 | 304.7 ± 7.0 |
| 1.75 | 625.1 ± 22.3 | 60.6 ± 2.3 | 3047.8 ± 73.9 | 3185.0 | 311.2 ± 7.3 |
| 2 | 693.0 ± 27.9 | 61.3 ± 1.8 | 3343.7 ± 80.1 | 3287.8 | 325.7 ± 9.2 |
| 2.5 | 888.7 ± 38.5 | 66.1 ± 1.8 | 3753.4 ± 91.8 | 3718.4 | 376.6 ± 8.5 |
| 3 | 1118.4 ± 48.0 | 80.3 ± 2.8 | 4462.1 ± 119.0 | 4467.4 | 459.8 ± 10.4 |

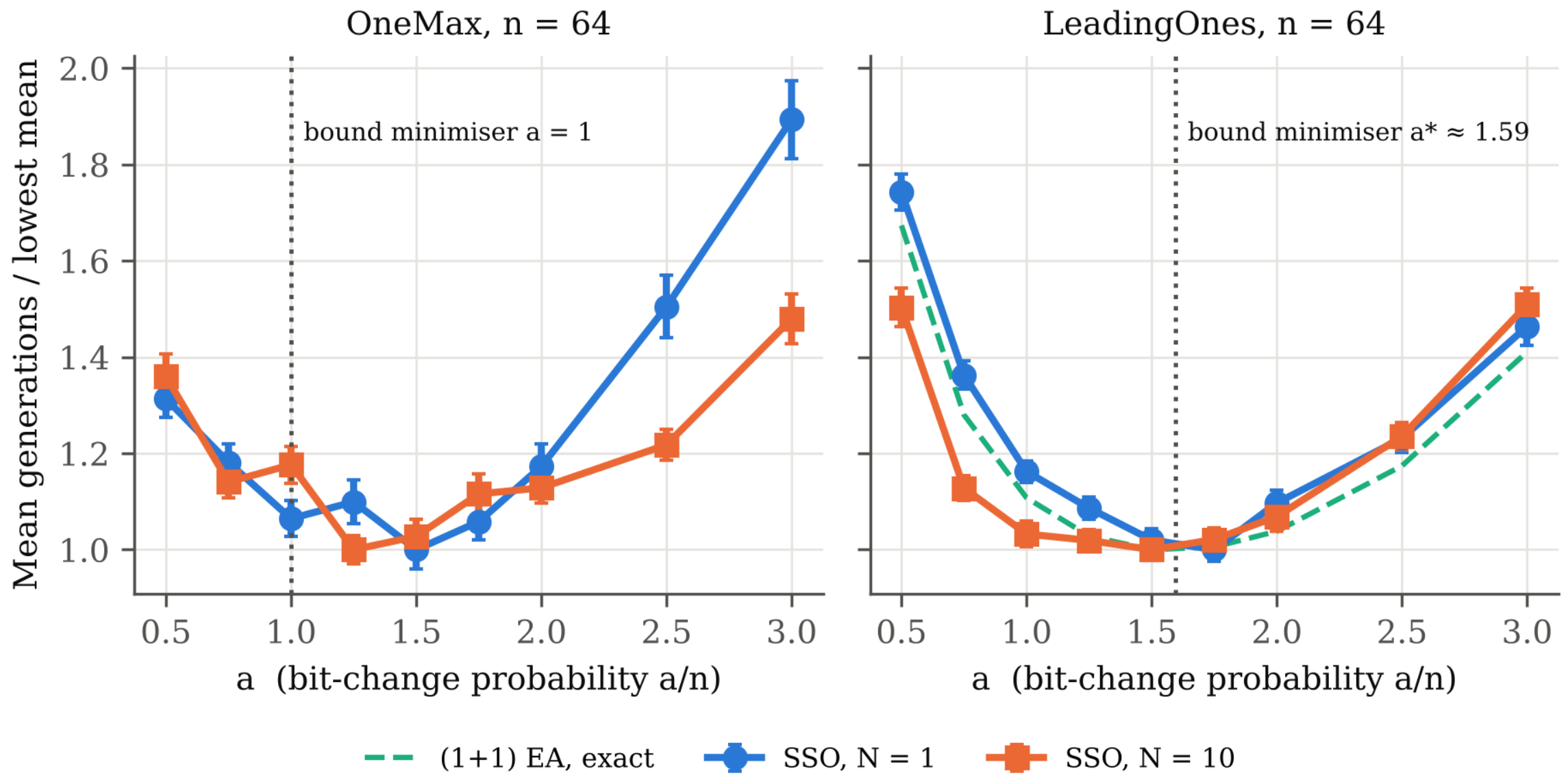

Fig. 4. Mean generations, normalised by the lowest mean of each curve, against the bit-change parameter a for n = 64 and $C_p = C_w$; error bars are one standard error. Dotted lines mark the bound minimisers of Corollary 3; the dashed curve is the exact (1+1) EA runtime on LeadingOnes [17].

### *F. The inertia weight*

Two experiments test Theorem 8. The first isolates the frozen-memory regime: the owner of g on OneMax with n = 64 and its personal best at distance d ∈ {1, 8} from the optimum. Its positions start either from stationarity, $x_0 \sim \pi$, which is the setting of Theorem 8(iii), or from $x_i = p_i$, the state after an improvement, which is a transient that the theorem does not cover. We count the updates until the first position better than $p_i$, for $c_w \in \{0, 0.3, 0.6, 0.9\}$ at a fixed redraw share $r \in \{2/n, 1/9\}$, with 4000 repetitions; without inertia successive positions are independent with law π, and the exact mean is $1/\pi(A)$ (Table VIII).

TABLE VIII. Frozen owner on OneMax (n = 64): mean number of updates until the first improvement (± standard error, 4000 repetitions) at a fixed redraw share, from stationary positions (Theorem 8(iii)) and from $x_i = p_i$

| Redraw share | d | Start | 1/π(A) | $c_w = 0$ | $c_w = 0.3$ | $c_w = 0.6$ | $c_w = 0.9$ |
|---|---|---|---|---|---|---|---|
| r = 2/n | 1 | $x_0 \sim \pi$ | 172.6 | 175.6 ± 2.8 | 209.0 ± 3.3 | 313.9 ± 5.0 | 1124.8 ± 17.9 |
| r = 2/n | 1 | $x_0 = p_i$ | 172.6 | 175.8 ± 2.8 | 218.0 ± 3.4 | 313.1 ± 5.0 | 1113.2 ± 18.1 |
| r = 2/n | 8 | $x_0 \sim \pi$ | 19.5 | 19.7 ± 0.3 | 23.7 ± 0.4 | 37.3 ± 0.6 | 127.9 ± 2.1 |
| r = 2/n | 8 | $x_0 = p_i$ | 19.5 | 19.4 ± 0.3 | 24.1 ± 0.4 | 36.2 ± 0.6 | 130.3 ± 2.1 |
| r = 1/9 | 1 | $x_0 \sim \pi$ | 659.4 | 657.2 ± 10.3 | 695.7 ± 10.7 | 773.5 ± 12.0 | 2063.0 ± 32.6 |
| r = 1/9 | 1 | $x_0 = p_i$ | 659.4 | 657.3 ± 10.2 | 659.8 ± 10.5 | 795.9 ± 12.6 | 2089.8 ± 34.7 |
| r = 1/9 | 8 | $x_0 \sim \pi$ | 38.4 | 37.9 ± 0.6 | 42.6 ± 0.7 | 54.1 ± 0.9 | 163.0 ± 2.6 |
| r = 1/9 | 8 | $x_0 = p_i$ | 38.4 | 37.6 ± 0.6 | 41.4 ± 0.7 | 53.6 ± 0.9 | 147.7 ± 2.6 |

The simulated means for $c_w = 0$ agree with the exact values. From stationary positions every $c_w > 0$ needs more updates than $1/\pi(A)$, as Theorem 8(iii) requires, up to 6.7 times as many at $c_w = 0.9$. From $x_i = p_i$ the means with inertia are also above $1/\pi(A)$ in all 12 cases, one of them within one standard error: on OneMax the transient of Theorem 8(ii), in which the effective mutation probability after an improvement starts at $c_r(m - 1)/m$ instead of $r(m - 1)/m$, does not offset the correlation, even when r = 1/9 is far above the end-game value 2/n, although Example 1 shows that transients can help on other landscapes.

The second experiment runs the full algorithm with N = 10, $c_g : c_p = 5 : 3$ and the redraw share fixed at r = 2/n or r = 1/9, for $c_w \in \{0, 0.1, \ldots, 0.7\}$. Only the inertia weight changes, so the stationary laws of Theorem 8(i) are the same for all $c_w$. Each setting has 200 runs, and each $c_w > 0$ is compared with $c_w = 0$ by two-sided Mann–Whitney U tests with Holm correction (Table IX, Fig. 5).

TABLE IX. Full SSO with a fixed redraw share $r = 2/n$: mean generations to the optimum ± standard error (200 runs, $N = 10$, $c_g : c_p = 5 : 3$); signs compare with $c_w = 0$

| $c_w$ | OneMax, n = 64 | LeadingOnes, n = 64 | $Jump_2$, n = 32 | $Jump_3$, n = 20 | $Jump_4$, n = 20 |
|---|---|---|---|---|---|
| 0 | 61 ± 2 | 332 ± 5 | 152 ± 12 | 630 ± 61 | 6,132 ± 640 |
| 0.1 | 61 ± 1 = | 353 ± 6 − | 155 ± 11 = | 481 ± 40 = | 4,634 ± 584 = |
| 0.2 | 69 ± 1 − | 376 ± 6 − | 149 ± 12 = | 410 ± 39 + | 6,134 ± 715 = |
| 0.3 | 76 ± 2 − | 411 ± 7 − | 130 ± 11 = | 416 ± 44 + | 4,446 ± 503 + |
| 0.4 | 82 ± 2 − | 459 ± 8 − | 142 ± 11 = | 439 ± 44 + | 3,991 ± 445 + |
| 0.5 | 94 ± 2 − | 540 ± 10 − | 163 ± 13 = | 379 ± 34 + | 2,982 ± 327 + |
| 0.6 | 118 ± 3 − | 656 ± 10 − | 139 ± 10 = | 386 ± 35 + | 3,386 ± 370 + |
| 0.7 | 146 ± 3 − | 841 ± 13 − | 201 ± 16 − | 486 ± 44 = | 2,855 ± 395 + |

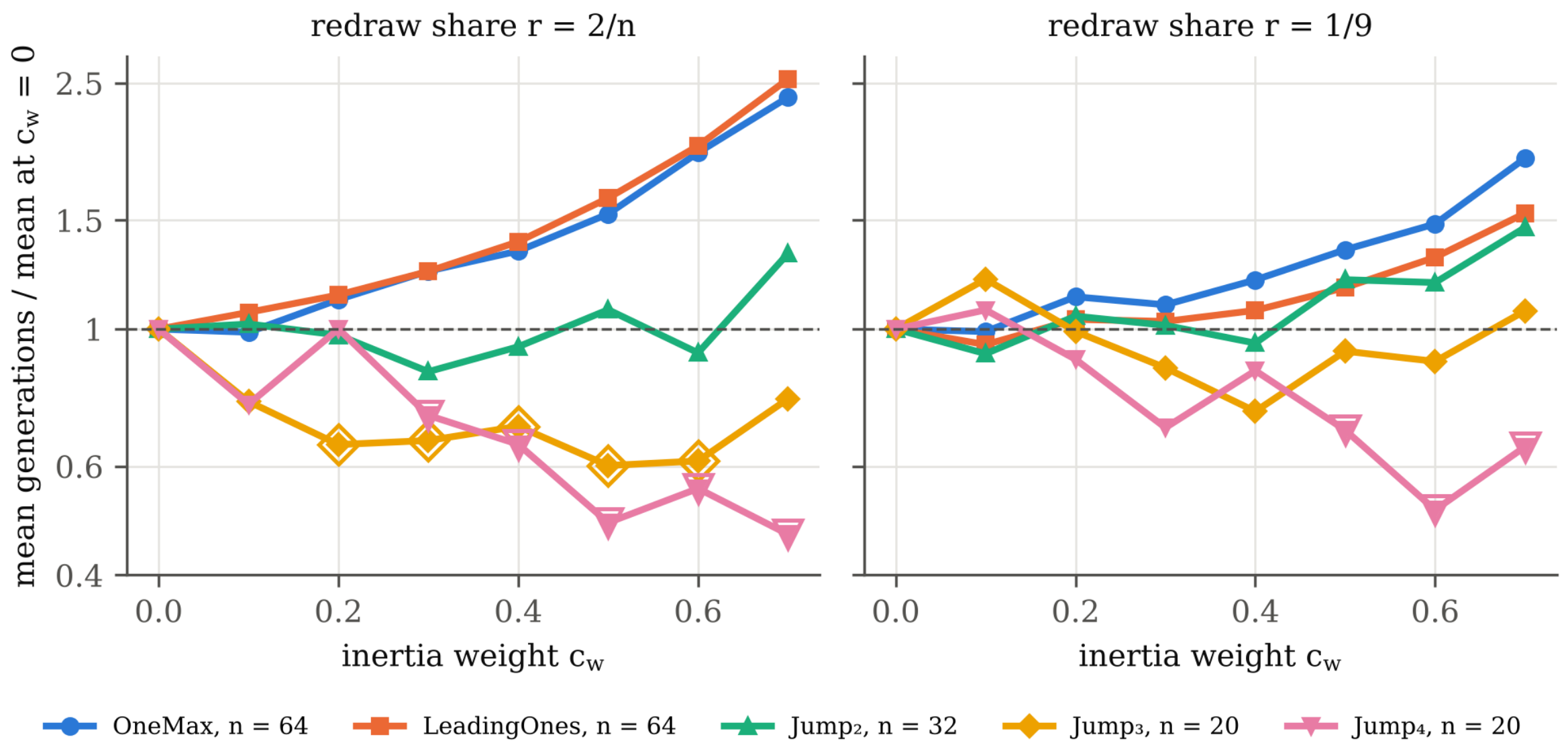

Fig. 5. Mean generations to the optimum, relative to $c_w = 0$, against the inertia weight at a fixed redraw share $r = 2/n$ (left) and $r = 1/9$ (right); open markers show a significant improvement over $c_w = 0$ (Holm-corrected Mann–Whitney tests, 200 runs).

On the tested OneMax and LeadingOnes settings no statistically significant improvement from positive inertia was detected at either redraw share, and $c_w = 0.7$ is significantly slower in all four cases, by a factor of up to 2.5. On $Jump_2$ no significant improvement was detected either. On $Jump_3$ and $Jump_4$, by contrast, moderate and large inertia weights are significantly faster: with $r = 2/n$ the mean falls by up to 40% on $Jump_3$ and 53% on $Jump_4$, and with $r = 1/9$ by up to 49% on $Jump_4$, while no inertia weight on $Jump_3$ with $r = 1/9$ differs significantly from $c_w = 0$. Theorem 8 rules out a gain from stationary positions with a fixed memory, but not a gain from the positions that occur in a run (Example 1) or from changes of the memory; Sections IX-G and IX-H separate these effects.

### *G. Coincidence and recombination in runs*

**Coincidence.** We recorded the class of Proposition 4 of every coordinate update in runs of the full update with $(c_g, c_p, c_w, c_r) = (0.5, 0.3, 0.1, 0.1)$: OneMax with $n = 64$ and $N = 10$ for 100 generations (50 runs), and the redundancy allocation problem of Section X-D with $N = 50$ for 5000 generations (10 runs at each of $W = 159$, 175 and 191). Table X lists the effective probabilities, the predicted and observed change probabilities and the share of each class, and Fig. 6 shows how the shares evolve.

TABLE X. Coincidence classes of $(g_j, p_{ij}, x_{ij})$ for $(c_g, c_p, c_w, c_r) = (0.5, 0.3, 0.1, 0.1)$: effective branch probabilities (Proposition 4), probability that the coordinate changes, and share of coordinate updates

| Class | Coincidence | Effective $(c_g, c_p, c_w, c_r)$ | $P(x'_{ij} \neq x_{ij})$ | m = 2: predicted / observed | m = 9: predicted / observed | Share (%): OneMax / RAP |
|---|---|---|---|---|---|---|
| A | none | $(c_g, c_p, c_w, c_r)$ | $c_g + c_p + c_r(m - 1)/m$ | — | 0.889 / 0.889 | 0 / 0.5 |
| B | $g_j = p_{ij} \neq x_{ij}$ | $(C_p, 0, c_w, c_r)$ | $C_p + c_r(m - 1)/m$ | 0.850 / 0.851 | 0.889 / 0.889 | 4.8 / 9.3 |
| C | $p_{ij} = x_{ij} \neq g_j$ | $(c_g, 0, c_p + c_w, c_r)$ | $c_g + c_r(m - 1)/m$ | 0.550 / 0.549 | 0.589 / 0.589 | 4.5 / 2.0 |
| D | $g_j = x_{ij} \neq p_{ij}$ | $(0, c_p, c_g + c_w, c_r)$ | $c_p + c_r(m - 1)/m$ | 0.350 / 0.349 | 0.389 / 0.389 | 2.7 / 3.0 |
| E | $g_j = p_{ij} = x_{ij}$ | $(0, 0, C_w, c_r)$ | $c_r(m - 1)/m$ | 0.050 / 0.050 | 0.089 / 0.089 | 87.9 / 85.3 |

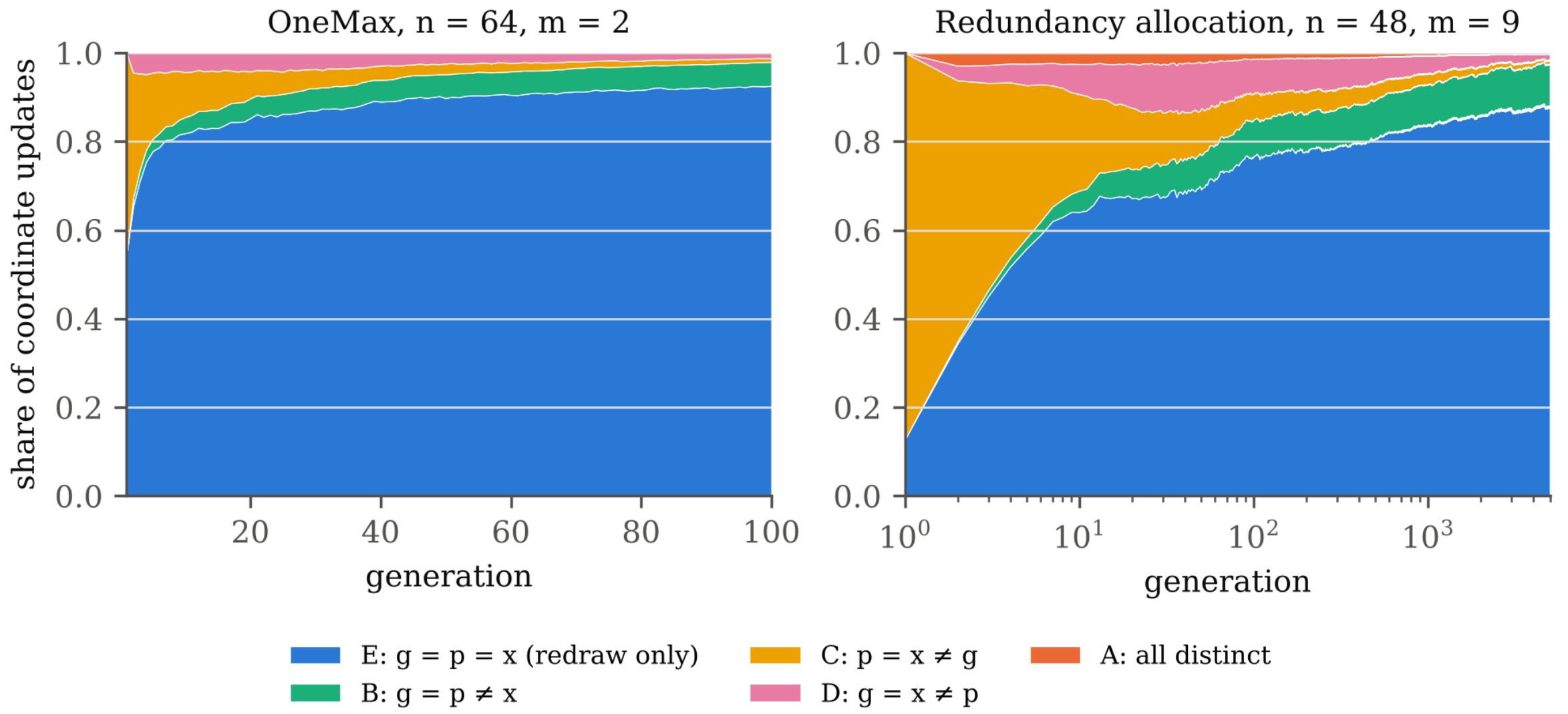

Fig. 6. Share of coordinate updates in each coincidence class during runs of the full update with $(c_g, c_p, c_w, c_r) = (0.5, 0.3, 0.1, 0.1)$: OneMax (50 runs, N = 10) and the redundancy allocation problem (W = 159, 175 and 191, 10 runs each, N = 50; logarithmic generation axis). In class E only the redraw branch can change the coordinate.

The observed change probabilities agree with Proposition 4 to within 0.001. Class E dominates: it accounts for 87.9% of all coordinate updates on OneMax and 85.3% on the redundancy allocation problem, and for 92.6% and 87.8% in the last generation, while class A, the only class in which all four branches act separately, accounts for 0.5% on the redundancy allocation problem and cannot occur on OneMax. Whole vectors rarely coincide after the first generations: $x_i = p_i$ holds for 2.8% of the solutions in the last OneMax generation and for at most 1.4% from generation 500 on in the redundancy allocation problem. Class E dominates because the references agree coordinate by coordinate, not because whole solutions coincide.

For most of a run, then, these parameters act on most coordinates as a uniform mutation that changes a coordinate with probability $c_r(m - 1)/m$, whatever the values of $c_g$, $c_p$ and $c_w$. This probability is 0.050 on OneMax and 0.089 on the redundancy allocation problem, so in the last generation the class-E coordinates alone contribute about 3.0 and 3.7 changed coordinates per new solution, roughly three to four times the single change at which Corollary 1 aims. The tuned redraw weight of Corollary 3 acts on exactly this quantity: it gives $n \cdot c_r(m - 1)/m = 1 - c_w = 0.9$ on both problems, which explains from a different angle why it improves these parameters in Section X.

**Recombination.** To test Proposition 5 and Corollary 4 we logged, in the generation in which the optimum of $\text{Jump}_k$ was first produced, the solution that produced it: the number $k_g$ of wrong coordinates of g; the number $I_{gp}$ of coordinates on which g and $p_i$ are both wrong; the number $I_+$ of coordinates on which no active branch holds the optimal value ($a_j = 0$ in Proposition 5), which therefore had to come from the redraw branch; and, from the random numbers of the update, the branch that actually produced the optimal value on each of the $k_g$ coordinates where g is wrong. The settings are the default $(c_g, c_p, c_w, c_r) = (0.5, 0.3, 0.1, 0.1)$, its tuned variant, and $r = 2/n$ with $c_g : c_p = 5 : 3$ and $c_w \in \{0, 0.3, 0.6\}$; N = 10 and 200 runs each (Table XI). These quantities are conditioned on success: they describe the offspring that crossed the gap, not the search that preceded it.

TABLE XI. The solution that first produced the optimum of $Jump_k$: mean wrong coordinates of g ($k_g$), of g and $p_i$ jointly ($I_{gp}$), coordinates that only the redraw branch could supply ($I_+$), share of crossings with $I_+ = 0$, and the branch that actually supplied the optimal value on the coordinates where g is wrong (200 runs)

| Problem | Setting | Mean generations | $k_g$ | $I_{gp}$ | $I_+$ | $I_+ = 0$ | Source on gap coordinates: $p_i$ / $x_i$ / redraw |
|---|---|---|---|---|---|---|---|
| $Jump_2$, n = 32 | Default | 71 | 2.00 | 0.84 | 0.74 | 39% | 46% / 8% / 46% |
| $Jump_2$, n = 32 | Tuned r | 158 | 2.00 | 1.34 | 1.18 | 16% | 26% / 11% / 63% |
| $Jump_3$, n = 20 | Default | 354 | 3.00 | 1.58 | 1.35 | 12% | 37% / 10% / 53% |
| $Jump_3$, n = 20 | Tuned r | 495 | 3.00 | 1.65 | 1.35 | 14% | 37% / 10% / 54% |
| $Jump_4$, n = 20 | Default | 3,563 | 4.00 | 2.10 | 1.83 | 8% | 36% / 10% / 55% |
| $Jump_4$, n = 20 | Tuned r | 6,134 | 4.00 | 2.28 | 1.96 | 4% | 35% / 9% / 56% |
| $Jump_3$, n = 20 | r = 2/n, $c_w$ = 0 | 521 | 3.00 | 1.74 | 1.74 | 6% | 38% / 0% / 63% |
| $Jump_3$, n = 20 | r = 2/n, $c_w$ = 0.3 | 422 | 3.00 | 1.60 | 1.15 | 24% | 30% / 26% / 45% |
| $Jump_3$, n = 20 | r = 2/n, $c_w$ = 0.6 | 357 | 3.00 | 1.32 | 0.55 | 52% | 20% / 57% / 23% |
| $Jump_4$, n = 20 | r = 2/n, $c_w$ = 0 | 7,123 | 4.00 | 2.31 | 2.31 | 2% | 38% / 0% / 62% |
| $Jump_4$, n = 20 | r = 2/n, $c_w$ = 0.3 | 4,679 | 4.00 | 2.07 | 1.47 | 15% | 32% / 24% / 44% |
| $Jump_4$, n = 20 | r = 2/n, $c_w$ = 0.6 | 3,024 | 4.00 | 1.73 | 0.71 | 41% | 23% / 56% / 21% |

In every setting g is at a local optimum when the gap is crossed ($k_g = k$). Under the default and tuned parameters only 37–59% of the k gap coordinates had to come from the redraw branch ($I_+/k$), and the branch logs show that 37–54% of the optimal values on these coordinates were in fact copied, mostly from $p_i$; realised redraws are somewhat more frequent than necessary ones, because a redraw sometimes produces a value that a copy could also have supplied. Without inertia the current position is not an active source, so $I_+ = I_{gp}$ and no gap coordinate is copied from $x_i$. With $c_w = 0.6$ the inertia branch supplied 57% and 56% of the gap coordinates on $Jump_3$ and $Jump_4$, $I_+$ fell from 1.74 to 0.55 and from 2.31 to 0.71, and $I_{gp}$ fell from 1.74 to 1.32 and from 2.31 to 1.73, so the memories were also more complementary. Because these statistics are conditioned on success, they show how the gap was crossed, not why inertia shortens the search; Section IX-H addresses that question.

### *H. Memory, transient and correlation effects*

To separate the effects of inertia listed in Remark 7, we ran SSO on $Jump_3$ and $Jump_4$ (n = 20, N = 10, $r = 2/n$, $c_g : c_p = 5 : 3$) with $c_w = 0$ and $c_w = 0.6$, 100 runs each. For the memory (g, P) of every generation we computed $h(g, P) = 1 - \prod_i(1 - \pi_i(x^*))$, where $\pi_i$ is the stationary law of solution i; by Theorem 8(i), h is the exact per-generation probability of producing x* for a frozen memory and positions drawn independently from their stationary laws, and it is the same function of the memory for every $c_w$ here. We saved the state in the first generation in which g was a local optimum and replayed each saved state five times under $c_w \in \{0, 0.3, 0.6\}$ with the same normalised probabilities, in three modes: with the memory frozen and the recorded positions; with the memory frozen and independent stationary positions; and with the memory free to change, which continues the run. Replays of the same states are compared with $c_w = 0$ by Wilcoxon signed-rank tests on the per-state means, with Holm correction (Table XII).

TABLE XII. Checkpoint replay: mean generations until x* is produced (± standard error over the 100 per-state means; 5 replays per state), by the inertia weight used to generate the saved states, the replay mode and the replay inertia weight, with the exact frozen-memory reference for $c_w = 0$; signs compare with replay $c_w = 0$ on the same states. The last row of each block gives the mean interaction contrast $\bar{D}$ of the text with its pointwise 95% exploratory bootstrap interval

| Problem | States generated with | Replay mode | Exact, $c_w$ = 0 | $c_w = 0$ | $c_w = 0.3$ | $c_w = 0.6$ |
|---|---|---|---|---|---|---|
| $Jump_3$ | $c_w = 0$ | Frozen memory, recorded positions | 1,052 | 1,086 ± 104 | 1,139 ± 108 = | 1,236 ± 113 = |
| $Jump_3$ | $c_w = 0$ | Frozen memory, stationary positions | 1,052 | 1,030 ± 98 | 1,096 ± 90 = | 1,217 ± 112 − |
| $Jump_3$ | $c_w = 0$ | Memory free (continued run) | — | 543 ± 34 | 545 ± 42 = | 606 ± 43 = |
| $Jump_3$ | $c_w = 0$ | Interaction $\bar{D}$ [pointwise 95% CI] | — | — | −51 [−243, 134] | −86 [−272, 97] |
| $Jump_3$ | $c_w = 0.6$ | Frozen memory, recorded positions | 703 | 672 ± 74 | 670 ± 67 = | 812 ± 82 − |
| $Jump_3$ | $c_w = 0.6$ | Frozen memory, stationary positions | 703 | 700 ± 79 | 737 ± 97 = | 876 ± 106 − |
| $Jump_3$ | $c_w = 0.6$ | Memory free (continued run) | — | 405 ± 33 | 363 ± 30 = | 357 ± 29 = |
| $Jump_3$ | $c_w = 0.6$ | Interaction $\bar{D}$ [pointwise 95% CI] | — | — | −40 [−157, 84] | −188 [−303, −80] |
| $Jump_4$ | $c_w = 0$ | Frozen memory, recorded positions | 8,476 | 8,941 ± 828 | 8,985 ± 921 = | 8,991 ± 774 = |
| $Jump_4$ | $c_w = 0$ | Frozen memory, stationary positions | 8,476 | 7,974 ± 761 | 7,755 ± 715 = | 10,729 ± 1,086 − |
| $Jump_4$ | $c_w = 0$ | Memory free (continued run) | — | 6,253 ± 432 | 4,992 ± 327 + | 5,236 ± 360 + |
| $Jump_4$ | $c_w = 0$ | Interaction $\bar{D}$ [pointwise 95% CI] | — | — | −1,305 [−3,177, 472] | −1,067 [−2,551, 408] |
| $Jump_4$ | $c_w = 0.6$ | Frozen memory, recorded positions | 5,039 | 4,928 ± 498 | 5,645 ± 790 = | 6,226 ± 813 − |
| $Jump_4$ | $c_w = 0.6$ | Frozen memory, stationary positions | 5,039 | 5,061 ± 529 | 5,214 ± 586 = | 5,579 ± 577 = |
| $Jump_4$ | $c_w = 0.6$ | Memory free (continued run) | — | 4,928 ± 331 | 3,935 ± 333 + | 3,735 ± 318 + |
| $Jump_4$ | $c_w = 0.6$ | Interaction $\bar{D}$ [pointwise 95% CI] | — | — | −1,709 [−2,996, −493] | −2,490 [−3,715, −1,321] |

Three results follow. First, the memories differ: when g reaches the gap, the mean of h for runs with $c_w$ = 0.6 is 1.9 times that for runs without inertia on $Jump_3$ and 1.6 times on $Jump_4$ (Mann–Whitney $p < 10^{-4}$), and over the remaining plateau phase it is 1.7 and 1.6 times as large; accordingly, every frozen replay is 28–48% faster from states generated with inertia, whatever the replay inertia weight. Second, with the memory frozen no inertia weight is significantly faster than $c_w = 0$, either from stationary positions, as Theorem 8(iii) predicts, or from the recorded positions, so the transient of the actual positions does not produce a measurable gain here; $c_w = 0.6$ is significantly slower in several cases, and the frozen replays without inertia agree with the exact reference $(1/M)\Sigma_s 1/h_s$, the mean of the exact expected times over the M = 100 saved states. Third, when the memory is free to change, inertia is significantly faster on $Jump_4$ from both sets of states (by 20% and 24%), whereas on $Jump_3$ no difference is detected.

A significant effect in one replay mode and a non-significant one in another do not by themselves show that the two effects differ. We therefore computed for each saved state s the contrast $D_s = [\bar{T}_{free}(s; c_w) - \bar{T}_{free}(s; 0)] - [\bar{T}_{frozen}(s; c_w) - \bar{T}_{frozen}(s; 0)]$, where $\bar{T}(s; c)$ is the mean number of generations in the five replays of state s with replay inertia c, using the frozen replays from the recorded positions, which start from the same positions as the continued runs. For its mean $\bar{D}$ we report a pointwise 95% exploratory percentile bootstrap interval, obtained from 10,000 resamples of the 100 saved states with replacement (seed 2026) and not adjusted for the eight comparisons; all eight estimates are listed in Table XII. They are computed from unrounded per-state means, so they can differ by one generation from contrasts of the rounded table entries. The mean of $D_s$ is negative in all eight combinations of problem, generating weight and replay weight, so inertia is relatively more favourable when the memory can change. The pointwise interval excludes zero for $Jump_4$ from states generated with inertia, with −1,709 [−2,996, −493] generations for replay $c_w = 0.3$ and −2,490 [−3,715, −1,321] for $c_w = 0.6$, and for $Jump_3$ from those states with $c_w = 0.6$ (−188 [−303, −80]); it contains zero in the other five combinations, for example −1,067 [−2,551, 408] for $Jump_4$ from states generated without inertia.

In summary, states generated with inertia had higher stationary success probabilities h and shorter frozen-memory replay times, no significant benefit from positive replay inertia was detected with the memory frozen, and inertia was relatively more favourable when the memory could change. Together these findings support a memory-mediated explanation of the gains of inertia on the tested Jump instances, without establishing that changes of memory are the only contributing mechanism.

## X. Experimental Study of the Original SSO

This section evaluates the original SSO under the conditions in which it is used: a population, default parameters with inertia, functions with fitness gaps, a range of redraw weights, and a constrained reliability problem with non-binary variables. Two references are used: the redraw rule of Corollary 3, which changes the redraw share while keeping the inertia weight and the ratio $c_g : c_p$, and the (1+1) EA, the mutation-only algorithm to which SSO reduces, apart from the acceptance of ties, for one solution without inertia (Remark 6(a)).

### *A. Experimental setup*

**Algorithms.** SSO with the full update of Algorithm 1 is run in two settings: S1 = $(c_g, c_p, c_w, c_r)$ = (0.25, 0.25, 0, 0.5), without inertia and with a large redraw weight, and S2 = (0.5, 0.3, 0.1, 0.1), the default setting of [5] converted with Table I. For each setting a tuned variant sets the redraw weight so that $r = m/((m - 1)n)$ (Corollary 3), keeping $c_w$ and the ratio $c_g : c_p$; the two copy probabilities are rescaled to preserve normalisation, so that for S2 with n = 64 the tuned vector is $(c_g, c_p, c_w, c_r)$ = (0.5449, 0.3270, 0.1, 0.0281). The baseline is the (1+1) EA [16] with standard mutation, which changes each coordinate with probability 1/n to a uniformly chosen different value and accepts the offspring if it is not worse.

**Benchmarks and protocol.** OneMax and LeadingOnes with n = 64, $Jump_2$ with n = 32 and $Jump_3$ with n = 20, where $Jump_k(x) = d$ if $d = 0$ or $d \geq k$ and $n + k - d$ if $0 < d < k$, with d the Hamming distance to x*. SSO uses N = 10 and stops after at most $10^5$ generations, that is, at most $10 \times (10^5 + 1) = 1{,}000{,}010$ evaluations including the initial population; the (1+1) EA stops after $10^6$ evaluations including its initial point. Each configuration has 50 runs. Costs are counted in evaluations, including the initial population, so that SSO and the (1+1) EA are compared on the same scale. The 10 additional evaluations available to SSO affect no comparison: the most expensive successful SSO run in Tables XIII and XIV used 658,980 evaluations.

**Statistics.** Two-sided Mann–Whitney U tests, with unsolved runs ranked last, compare each tuned variant with the default of the same setting (Holm correction over the eight problem–setting pairs) and each SSO configuration with the (1+1) EA (Holm correction over the four configurations of each problem). “+” and “−” mark significantly fewer or more evaluations at the 5% level, and “=” no significant difference. Code, seeds and per-run results are in the supplementary material.

## *B. Benchmark functions*

TABLE XIII. Mean evaluations to the optimum ± standard error over solved runs (50 runs, SSO with N = 10). The first sign compares with the (1+1) EA; the second, for tuned variants, with the default of the same setting

| Problem | (1+1) EA | SSO, S1 | SSO, S1, tuned r | SSO, S2 | SSO, S2, tuned r |
|---|---|---|---|---|---|
| OneMax, n = 64 | 654 ± 38 | 0% solved (−) | 595 ± 25 (=, +) | 988 ± 52 (−) | 639 ± 27 (=, +) |
| LeadingOnes, n = 64 | 3,551 ± 109 | 0% solved (−) | 2,859 ± 83 (+, +) | 6,076 ± 280 (−) | 3,405 ± 93 (=, +) |
| $Jump_2$, n = 32 | 2,931 ± 380 | 64,986 ± 9,040 (−) | 1,026 ± 176 (+, +) | 1,092 ± 127 (+) | 1,496 ± 244 (+, =) |
| $Jump_3$, n = 20 | 19,924 ± 2,804 | 5,072 ± 604 (+) | 3,072 ± 701 (+, +) | 3,191 ± 661 (+) | 4,233 ± 829 (+, =) |

Table XIII summarises the runs. The tuned redraw weight is significantly better than the default in 6 of the eight problem–setting pairs, and no statistically significant deterioration was detected in the other 2, although on $Jump_2$ and $Jump_3$ under S2 its sample means are higher. With the 200 runs of Table XIV the $Jump_2$ difference is significant (see below); for n = 20 the two weights are close ($c_r$ = 0.09 against 0.1), and no difference is detected on $Jump_3$. Without tuning, S1 fails on OneMax and LeadingOnes, where its redraw weight of 0.5 changes about 16 of the 64 coordinates of a converged solution, so that by Proposition 2(iii) a run reaches the optimum within the budget with probability less than 1.01%; S2 needs 1.5 and 1.7 times as many evaluations as the (1+1) EA. With the tuned redraw weight SSO showed no statistically significant difference from the (1+1) EA on OneMax in both settings and on LeadingOnes under S2, and was significantly faster on LeadingOnes under S1. On the Jump functions the picture reverses: every SSO configuration except S1 on $Jump_2$ needs significantly fewer evaluations than the (1+1) EA with rate 1/n, the default S2 by factors of 2.7 on $Jump_2$ and 6.2 on $Jump_3$. This comparison is tied to the rate 1/n, which is far from the rate k/n that maximises the final-jump probability $p^k(1 - p)^{n-k}$: for $Jump_3$ with n = 20 the expected waiting time for the final jump is about 19,100 evaluations at p = 1/n and 4,700 at p = 3/n. The controls below remove this dependence.

**Mechanism-matched controls.** To see what the advantage on Jump functions is relative to, we ran four controls with 200 runs each and a cap of $10^6$ evaluations: the (1+1) EA with rate 1/n and with rate k/n; a population-matched (10+10) EA with rate 1/n, in which each parent creates one offspring and the best 10 of the 20 survive; and a (μ+1) GA with μ = 10, uniform crossover of two random parents with probability 1/2, mutation rate 1/n and replacement of a worst individual [14]. The (1+1) EAs accept an offspring that is not worse; the (10+10) EA ranks parents and offspring together by fitness, prefers offspring among equal fitness and breaks the remaining ties at random; the GA draws both parents uniformly with replacement, so the same parent can be drawn twice, copies the first parent when no crossover is applied, allows duplicate individuals, and replaces a uniformly chosen worst individual if the offspring is not worse. The SSO rows use the 200 runs per setting of Section IX-F: the default S2, its tuned variant, and r = 2/n with $c_w$ = 0.6. Table XIV reports the expected running time ERT, defined as the total number of evaluations of all runs, with unsuccessful runs charged the cap, divided by the number of successful runs; it equals the mean when every run succeeds.

TABLE XIV. Jump functions: expected running time ERT in evaluations (200 runs; cap $10^6$ evaluations for the controls and $10^5$ generations, at most 1,000,010 evaluations, for SSO, all of whose runs succeeded), with the success rate where it is below 100%

| Algorithm | $Jump_2$, n = 32 | $Jump_3$, n = 20 | $Jump_4$, n = 20 |
|---|---|---|---|
| (1+1) EA, rate 1/n | 2,840 | 19,496 | 392,111 (92% solved) |
| (1+1) EA, rate k/n | 1,791 | 4,915 | 23,548 |
| (10+10) EA, rate 1/n | 3,413 | 19,659 | 354,614 (95% solved) |
| (μ+1) GA, uniform crossover | 757 | 2,740 | 25,021 |
| SSO, default (S2) | 747 | 4,464 | 44,291 |
| SSO, tuned redraw weight | 1,559 | 4,821 | 46,351 |

| Algorithm | $Jump_2$, n = 32 | $Jump_3$, n = 20 | $Jump_4$, n = 20 |
|---|---|---|---|
| SSO, r = 2/n, $c_w$ = 0.6 | 1,402 | 3,865 | 33,874 |

SSO with the default parameters needs 3.8–8.9 times fewer evaluations than the (1+1) EA with rate 1/n and 4.4–8.0 times fewer than the population-matched (10+10) EA on all three functions (Holm-corrected Mann–Whitney tests, $p < 0.05$). Raising the mutation rate of the (1+1) EA to k/n closes most of this gap: on $Jump_4$ the (1+1) EA with rate k/n is significantly faster than the default SSO, and SSO with $c_w = 0.6$ does not differ significantly from it. No SSO configuration was significantly faster than the (μ+1) GA with uniform crossover on any of the three functions, and the GA was significantly faster than the default SSO on $Jump_3$. The additional controls therefore substantially reduce the advantage observed against the standard-rate baseline: performance depends on the instance and the parameter setting, and the experiments establish no general superiority of SSO over rate-adjusted mutation or uniform crossover.

**Cost of the end-game redraw weight.** The SSO rows of Table XIV also compare the default S2 ($c_r = 0.1$, redraw share r = 0.1/0.9) with its tuned variant, which sets r = 2/n and keeps $c_w = 0.1$, so that $c_r = 0.9 \cdot 2/n$. For n = 20 this gives $c_r = 0.09$, close to the default, and no difference was detected on $Jump_3$ or $Jump_4$ (p = 0.39 and 0.94). On $Jump_2$ with n = 32 it gives $c_r = 0.05625$, which is significantly slower: the median cost is 1,050 evaluations against 530 (two-sided Mann–Whitney test on the 200 + 200 per-run records, $p < 10^{-9}$; the records are the default and tuned entries with $c_w = 0.1$ in the sweep of expO.json in the package, with cost 10(G + 1) for a run of G generations). This is the trade-off of Remark 3. From a state with $g = p_i = x_i$ at the local optimum, crossing the two-bit gap needs two simultaneous changes, and the direct-hit probability $(c_r/2)^2(1 - c_r/2)^{30}$ is 1.60 times larger at $c_r = 0.1$ than at 0.05625; it is maximised at $c_r = 4/n = 0.125$. In this coincident-reference state the smaller redraw weight lowers the probability of crossing the two-coordinate gap directly, and the full runs show a corresponding slowdown on the tested $Jump_2$ instance.

### *C. Robustness to the redraw weight*

The redraw weight is the parameter that Corollary 3 asks to scale with 1/n, and in practice it is rarely tuned to the dimension. To measure the sensitivity of SSO to it, $c_r$ was varied over {0.02, 0.05, 0.1, 0.2, 0.3, 0.5} with $c_w = 0.1$ and $c_g : c_p = 5 : 3$, on OneMax and LeadingOnes with n = 64, N = 10 and 50 runs capped at $10^5$ generations (Table XV); the tuned value for n = 64 is $c_r = 0.028$. Because some settings fail in most runs, Table XV reports the success rate with a 95% Wilson interval and the ERT in generations instead of means over successful runs.

TABLE XV. Success rate (95% Wilson interval) and expected running time ERT in generations against the redraw weight (n = 64, N = 10, $c_w = 0.1$, $c_g : c_p = 5 : 3$, 50 runs, cap $10^5$ generations); ∞ means that no run succeeded, so the empirical ERT is undefined, not that the expected hitting time is infinite

| $c_r$ | OneMax: success | OneMax: ERT | LeadingOnes: success | LeadingOnes: ERT |
|---|---|---|---|---|
| 0.02 | 50/50 (93–100%) | 79 | 50/50 (93–100%) | 443 |
| 0.05 | 50/50 (93–100%) | 65 | 50/50 (93–100%) | 323 |
| 0.1 | 50/50 (93–100%) | 108 | 50/50 (93–100%) | 598 |
| 0.2 | 50/50 (93–100%) | 1,456 | 50/50 (93–100%) | 7,250 |
| 0.3 | 48/50 (87–99%) | 46,141 | 2/50 (1–13%) | $2.47\times10^6$ |
| 0.5 | 0/50 (0–7%) | ∞ | 0/50 (0–7%) | ∞ |

Every run is solved for $c_r \leq 0.2$, and the smallest ERTs are at 0.05 and 0.02, the two grid points around the tuned value. Beyond them the cost rises steeply: at $c_r = 0.2$ the ERT is about 22 times that at 0.05; at $c_r = 0.3$, 48 of 50 OneMax runs and 2 of 50 LeadingOnes runs reach the optimum within $10^5$ generations, which gives ERTs of 46,141 and $2.47\times10^6$ generations; and at $c_r = 0.5$ no run does, consistent with Proposition 2(iii), by which a run reaches the optimum within $10^5$ generations with probability less than 1.01%. In class E of Proposition 4 the redraw branch alone moves a coordinate, so $c_r = 0.3$ changes about $64 \times 0.3/2 \approx 10$ coordinates of a converged solution, and improvements near the optimum become exponentially rare (Corollary 1).

### *D. Case study: redundancy allocation*

The series–parallel redundancy allocation problem of Fyffe et al. [22] is a standard benchmark for SSO and other metaheuristics in reliability engineering. We use its mixed-component form [23]: 14 subsystems in series, each with three or four component types with given reliability, cost and weight; subsystem i receives $x_{ij} \geq 0$ components of type j, with $1 \leq \Sigma_j x_{ij} \leq 8$, and the system reliability $\prod_i[1 - \prod_j u_{ij}]$, with $u_{ij}$ the $x_{ij}$-th power of $1 - r_{ij}$, is maximised subject to a cost limit of 130 and a weight limit W, for the 33 limits W = 159, …, 191. The decision vector has n = 48 variables with values {0, …, 8}, so m = 9. Constraints are handled by the feasibility rule: a feasible point scores $1 - R(x)$ and an infeasible one 1 plus its normalised constraint violation, so every feasible point is preferred to every infeasible one. This changes f but not the kernel, so the kernel results apply (Theorems 1–3 and 8, Propositions 1, 4 and 5); the runtime bounds of Section VI and Corollary 4 are specific to their objective functions and do not transfer.

Exact optima were computed by dynamic programming over (cost, weight); they agree with the exact values reported in [24] to within $10^{-8}$ for all 33 instances. SSO used setting S2 and its tuned variant ($c_r = 0.021$, from $r = 9/(8 \cdot 48)$), with N = 50, a uniformly random initial population and 5000 further generations, that is, 50 × (5000 + 1) = 250,050 evaluations; the (1+1) EA was stopped after 250,000 evaluations including its initial point, so SSO had 50 more evaluations (0.02%, one generation of the swarm). Each algorithm had 30 runs per instance, and performance is measured by the relative gap $(R^* - R)/R^*$ of the best feasible solution found. For each instance the tuned SSO and the (1+1) EA are compared with the default SSO, and the tuned SSO with the (1+1) EA, by Mann–Whitney tests with Holm correction.

TABLE XVI. Redundancy allocation over the 33 instances: relative gap to the exact optimum (30 runs per instance; 250,050 evaluations for SSO and 250,000 for the (1+1) EA)

| Algorithm | Mean gap (%) | Median gap (%) | Best gap (%) | Feasible runs | Better / equal / worse than default SSO |
|---|---|---|---|---|---|
| SSO, default (S2) | 2.47 | 2.32 | 0.611 | 100% | — |
| SSO, tuned redraw weight | 2.19 | 1.83 | 0.500 | 100% | 19 / 14 / 0 |
| (1+1) EA | 2.35 | 1.76 | 0.385 | 100% | 17 / 16 / 0 |

Table XVI summarises the results. No run of any algorithm reached the exact optimum within the budget, and all runs ended with a feasible solution; the smallest gap, 0.385%, was reached by the (1+1) EA. The tuned redraw weight lowers the mean gap of SSO from 2.47% to 2.19% and is significantly better than the default on 19 instances and worse on none. The (1+1) EA, with a mean gap of 2.35% and the lowest median gap, is better than the default SSO on 17 instances and worse on none, and it does not differ significantly from the tuned SSO on any instance. The gaps of all three fall as the weight limit is relaxed (Fig. 7). On this constrained problem no significant difference between the tuned SSO and mutation-based search with the standard rate was detected on any instance, and the advantage seen on Jump functions does not appear.

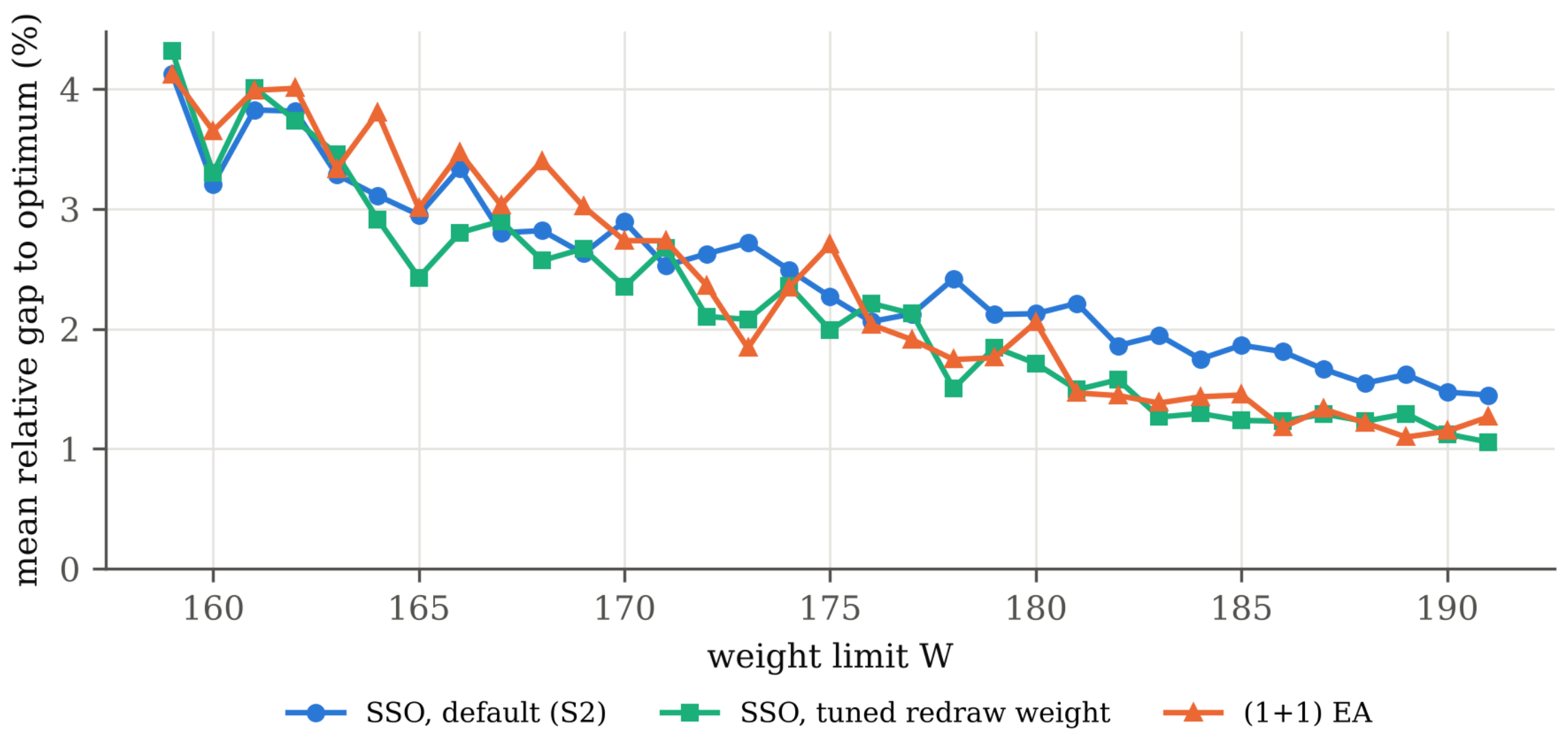

Fig. 7. Redundancy allocation: mean relative gap to the exact optimum against the weight limit W for the default SSO, the SSO with the tuned redraw weight and the (1+1) EA (30 runs per instance).

### *E. Summary of the experimental findings*

(i) The redraw weight matters most. The default weights of S1 and S2 either fail on OneMax and LeadingOnes or need up to 1.7 times as many evaluations as the (1+1) EA, and the rule $r = m/((m-1)n)$ of Corollary 3 removes this gap while keeping $c_w$ and the ratio $c_g : c_p$; no statistically significant deterioration was detected on any benchmark–setting pair of Table XIII or on any redundancy allocation instance, but with 200 runs the tuned weight was significantly slower than $c_r = 0.1$ on $\text{Jump}_2$, consistent with the one-step trade-off of Remark 3. (ii) On Jump functions SSO needs several times fewer evaluations than mutation with the standard rate 1/n, and its copy branches supply a large share of the gap coordinates (Proposition 5, Corollary 4, Table XI); against the (1+1) EA with rate k/n and the (μ+1) GA with uniform crossover no consistent advantage was established. (iii) On the tested OneMax and LeadingOnes settings positive inertia gave no significant improvement and large values were slower; on $\text{Jump}_3$ and $\text{Jump}_4$ it shortened the search, and checkpoint replay supports a memory-mediated explanation. These results suggest setting $c_w$ to zero or a small value on unimodal problems and treating it as a diversity parameter otherwise. (iv) On the constrained reliability problem the tuned SSO is significantly better than the default SSO on 19 of the 33 instances, and no significant difference from the (1+1) EA was detected.

## XI. Discussion

### *A. Design guidelines*

Table XVII collects the design guidelines that follow from the analysis and the experiments.

TABLE XVII. Design guidelines for discrete SSO

| Quantity | Role | Guideline | Source |
|---|---|---|---|
| $c_r = 1 - C_w$ | Global reachability; repair of wrong coordinates; sole source of change where $g_j = p_{ij} = x_{ij}$ | Keep > 0; a $c_r$ fixed during the run must be O((log n)/n) for a polynomial expected runtime to a unique optimum (necessary, not sufficient; Prop. 2(iii)); end-game value $m/((m-1)n)$ at distance one; on the tested OneMax and LeadingOnes settings ($n = 64$, $10^5$ generations), $c_r \geq 0.2$ had more than 20 times the ERT of $c_r = 0.05$ and no run with $c_r = 0.5$ succeeded; a gap of k coordinates favours $km/((m-1)n)$ (Rem. 3), and on $\text{Jump}_2$ the end-game value was slower than 0.1 | Lemma 1; Prop. 2; Thm. 4; Cor. 1; Prop. 4; Sec. X-C |

| Quantity | Role | Guideline | Source |
|---|---|---|---|
| $r = c_r/(1 - c_w)$ | Redraw share; controls the runtime bounds and the stationary sampling law | OneMax bound minimised at $m/((m - 1)n)$; LeadingOnes bound asymptotically at $\approx 1.59m/((m - 1)n)$ | Thms. 6–8; Cor. 3 |
| $c_w = C_w - C_p$ | Inertia: correlation between successive positions | Cannot speed up sampling from stationary positions for a fixed memory (other starts can differ, Ex. 1); no gain detected on the tested OneMax and LeadingOnes settings; on the tested Jump functions moderate values helped, consistent with a memory-mediated mechanism | Thms. 1, 8; Ex. 1; Sec. IX-F to IX-H |
| $c_g = C_g$ | Preserves correct coordinates near the optimum | Larger $c_g$ raises the gBest-guided lower bound on one-step success near the optimum | Thm. 4 |
| $c_p = C_p - C_g$ | Pull towards the personal best; second source for recombination | Acts on its own only where $p_{ij}$ differs from $g_j$ and $x_{ij}$; keep $c_g$, $c_p > 0$ so that complementary memories can be recombined | Props. 4–5; Cor. 4 |
| Mutation type (bits) | Uniform redraw or flip | For flip, $C_p < C_w < 1$ guarantees the one-step minorisation; decay rate $\lvert c_w - c_r \rvert^t$ | Prop. 3; Thm. 5 |

### *B. Limitations and extensions*

- **From convergence to runtime.** Theorem 3 is a worst-case bound; it is attained by needle functions without inertia (Proposition 2(ii)) and is loose on landscapes with fitness information (Table IV). Section VI shows that fitness levels combined with the owner lemma give polynomial bounds within a factor of 13–78 of the observed means (Table VI), and within a factor of about 2 of the exact runtime when $N = 1$ without inertia. Removing the $\log n$ factor for $c_w > 0$, proving a speed-up in N, extending the analysis to linear functions by drift analysis [9], and bounding the probability that complementary memories arise on $\text{Jump}_k$, which Corollary 4 leaves open, are natural next steps. Together, Proposition 2(iii) and the $\text{Jump}_2$ result of Section X-B indicate that a fixed redraw weight trades the two tasks against each other: it must be $O((\log n)/n)$ for a unique optimum to be reached in polynomial expected time, the end-game value repairs single coordinates fastest, and a gap of k coordinates favours a weight k times larger (Remark 3). A weight that decreases during the run, as the temperature does in simulated annealing [25], or one drawn anew in each generation from a heavy-tailed distribution, as in the fast genetic algorithms of [26], may combine the two; such schedules are not covered by the present analysis.
- **Inertia beyond stationarity.** Theorem 8 compares inertia with independent sampling only from stationary positions, and Example 1 shows that no such comparison holds from arbitrary positions. A bound on the transient effect of the positions that arise in a run, and on how inertia changes the memories the swarm builds, is open.
- **Asynchronous update.** If g is updated after each solution, each solution still satisfies Lemma 1 given the current state, so Theorems 2–3 hold with the same bound by the chain rule. The coupling of Lemma 4 would need to be redone for that schedule.
- **Mixed and continuous variables.** The kernel is a product over variables, so the minorisation constants multiply across variable types. The continuous case, where the copy branches become perturbations with a decreasing step size, is analysed in [18].
- **Partial updates.** Variants that update only some variables per generation change the redraw share and the anchor of the update, and are not covered by the results above; they are left to a separate study.
- **Threats to validity.** The binary benchmarks are of moderate size, the constrained study uses one problem family with uniform initialisation and a fixed budget, and the Jump comparisons use a small set of mechanism-matched controls. The experiments are designed to test the mechanisms identified by the theory, not to rank SSO against the many algorithms proposed for these problems; statements that no significant difference was detected are not claims of equivalence.

- **Reproducibility.** The transition-matrix generator, the solvers for the exact chains, the simulation code with its random seeds, the per-run results and the scripts that produce every table and figure are provided as supplementary material. A single script, reproduce.sh, reruns the experiments behind Tables XI–XVI, recomputes the statistics and the table entries, and checks that every regenerated file is identical to the stored records. The package also records the software versions used and SHA-256 checksums of all its files.

## XII. Conclusion

For the SSO update rule on a finite domain, the frozen-memory coordinate chain admits an exact spectral analysis: its kernel is a rank-one perturbation of a scaled identity, its stationary law is explicit, and its distance to stationarity decays exactly as $c_w^t$. The elitist chain converges with a geometric hitting-time bound that is attained by needle functions without inertia, and a gBest-guided bound gives the end-game redraw weight $m/((m-1)n)$; conversely, from a non-optimal initial global best any redraw weight held fixed during the run needs at least $\exp(nc_r/2)$ expected evaluations to reach a unique optimum, so this order cannot be exceeded by more than a log n factor. For a complement-mutation variant, $C_p < C_w < 1$ guarantees a one-step minorisation. Because the global best is always the personal best of some solution, an explicit coupling and the exact decay rate give runtime bounds for every population size: with a redraw share of order 1/n, $O(n \log n)$ and $O(n^2)$ expected generations on OneMax and LeadingOnes without inertia, and within a log n factor of these for any fixed inertia weight. For a fixed memory, inertia changes neither the stationary sampling law after normalisation nor, therefore, the stationary distribution of the distance between a solution and its memory, and from stationary positions it cannot speed up the first entry into any set, because the kernel is reversible with nonnegative spectrum; from other positions it can, as an exact example shows. The three reference values combine in two ways: where they agree, the branches merge and only $c_r$ acts, which covers most coordinate updates in practice; where they disagree, the copy branches recombine the memories, and a redraw is forced only where no active branch holds the target value. The experiments support these results within their scope. The redraw rule of the analysis improved the default parameters on 6 of 8 benchmark–setting pairs and 19 of 33 redundancy allocation instances; it was slower than $c_r = 0.1$ on $\text{Jump}_2$, where two coordinates must change at once, consistent with the one-step trade-off of Remark 3. On Jump functions, branch logs show that copying supplied a large share of the gap coordinates, and SSO needed 3.8 to 8.9 times fewer evaluations than mutation with the standard rate, but no consistent advantage over rate-adjusted mutation or a $(\mu+1)$ GA with uniform crossover was established. Inertia slowed the tested OneMax and LeadingOnes runs, and checkpoint replay supports a memory-mediated explanation of its gains on wider gaps.

## CRediT authorship contribution statement

**Wei-Chang Yeh:** Conceptualization, Methodology, Formal analysis, Software, Validation, Investigation, Data curation, Visualization, Writing – original draft, Writing – review & editing, Funding acquisition.

## Declaration of competing interest

The author is a named inventor of Taiwan patent I528298, "Simplified Swarm Optimization Method and System Thereof", which relates to the algorithm studied in this article.

## Funding

This work was supported by the National Science and Technology Council, Taiwan (grant numbers NSTC 114-2221-E-007-123-MY3 and NSTC 113-2221-E-007-117-MY3). The funding source had no involvement in the study design, the analysis and interpretation of the results, the writing of the article or the decision to submit it for publication.

## Data availability

The code, the per-run records, the saved replay states and the reproduction workflow (SSO_code_data_v12.2) are openly available in Zenodo, doi: 10.5281/zenodo.24153497 [27].

## Declaration of generative AI and AI-assisted technologies in the manuscript preparation process

During the preparation of this work, the author used Claude (Anthropic) and ChatGPT (OpenAI) in order to assist in writing, editing, and reviewing the manuscript text, code, derivations, statements, and results. After using these tools, the author reviewed and edited the content as needed and takes full responsibility for the content of the published article.